\documentclass[10pt,journal,compsoc]{IEEEtran}
\ifCLASSOPTIONcompsoc
  \usepackage[nocompress]{cite}
\else
  \usepackage{cite}
\fi
\ifCLASSINFOpdf
  \usepackage[pdftex]{graphicx}
\else
\fi
\usepackage{amsmath}
\usepackage{url}

\usepackage{booktabs}
\usepackage[utf8]{inputenc}
\usepackage{amssymb}
\usepackage{multirow}
\usepackage[dvinames,dvipsnames]{xcolor}
\usepackage{wrapfig}
\usepackage{algorithm2e,algorithmic}

\usepackage{scalerel,stackengine}
\stackMath
\newcommand\reallywidehat[1]{%
\savestack{\tmpbox}{\stretchto{%
  \scaleto{%
    \scalerel*[\widthof{\ensuremath{#1}}]{\kern-.6pt\bigwedge\kern-.6pt}%
    {\rule[-\textheight/2]{1ex}{\textheight}}
  }{\textheight}%
}{0.5ex}}%
\stackon[1pt]{#1}{\tmpbox}%
}
\usepackage{bm}

\newcommand{\CRCnote}[1]{}

\begin{document}
%
\title{Auto-Pytorch: Multi-Fidelity MetaLearning for Efficient and Robust AutoDL}
%
%
%
%

\author{Lucas Zimmer,~\IEEEmembership{}
        Marius Lindauer~\IEEEmembership{}
        and Frank Hutter~\IEEEmembership{}
\IEEEcompsocitemizethanks{\IEEEcompsocthanksitem L. Zimmer is with the Institute of Computer Science, University of Freiburg, Germany. E-mail: zimmerl@cs.uni-freiburg.de
\IEEEcompsocthanksitem M. Lindauer is with the Institute of Information Processing,  Leibniz University Hannover. 
E-mail: lindauer@tnt.uni-hannover.de 
\IEEEcompsocthanksitem F. Hutter is with the University of Freiburg and the Bosch Center for Artificial Intelligence.
E-mail: fh@cs.uni-freiburg.de}
}

%
%

\markboth{}%
{Zimmer \MakeLowercase{\textit{et al.}}: Auto-PyTorch}
%



\IEEEtitleabstractindextext{%

\begin{abstract}

While early AutoML frameworks focused on optimizing traditional ML pipelines and their hyperparameters, a recent trend in AutoML is to focus on neural architecture search. In this paper, we introduce Auto-PyTorch, which brings together the best of these two worlds by jointly and robustly optimizing the network architecture and the training hyperparameters to enable fully automated deep learning (AutoDL). Auto-PyTorch achieves state-of-the-art performance on several tabular benchmarks by combining multi-fidelity optimization with portfolio construction for warmstarting and ensembling of deep neural networks (DNNs) and common baselines for tabular data. To thoroughly study our assumptions on how to design such an AutoDL system, we additionally introduce a new benchmark on learning curves for DNNs, dubbed LCBench, and run extensive ablation studies of the full Auto-PyTorch on typical AutoML benchmarks, eventually showing that Auto-PyTorch performs better than several state-of-the-art competitors.
\end{abstract}


\begin{IEEEkeywords}
Machine learning, Deep learning, Automated machine learning, Hyperparameter optimization, Neural architecture search, Multi-fidelity optimization, Meta-learning 
\end{IEEEkeywords}}

\maketitle

\IEEEdisplaynontitleabstractindextext

%
\IEEEpeerreviewmaketitle


\IEEEraisesectionheading{\section{Introduction}\label{sec:introduction}}

%
%
%
%

In recent years, machine learning (ML) has experienced enormous growth in popularity as the last decade of research efforts has led to high performances on many tasks. Nevertheless, manual ML design is a tedious task, slowing down new applications of ML. Consequently, the field of automated machine learning (AutoML)~\cite{automl_book} has emerged to support researchers and developers in developing high-performance ML pipelines. Popular AutoML systems, such as Auto-WEKA~\cite{thornton-kdd13a}, hyperopt-sklearn~\cite{komer-automl14a}, TPOT~\cite{olson-gecco16a},  auto-sklearn~\cite{feurer-aaai15a}
or Auto-Keras~\cite{jin-kdd19a}
are readily available and allow for using ML with a few lines of code.

While initially most AutoML frameworks focused on traditional machine learning algorithms, deep learning (DL) has become far more popular over the last few years due to its strong performance and its ability to automatically learn useful representations from data. While choosing the correct hyperparameters for a task is of major importance for DL~\cite{automl_book}, the optimal choice of neural architecture also varies between tasks and datasets~\cite{zoph-iclr17a}.
This has led to the recent rise of neural architecture search (NAS) methods~\cite{elsken-jmlr19a}. Since there are interaction effects between the best architecture and the best hyperparameter configuration for it, AutoDL systems have to jointly optimize both~\cite{zela-automl18a}. 

Traditional approaches for AutoML, such as blackbox Bayesian optimization, evolutionary strategies or reinforcement learning, perform well for cheaper ML pipelines but hardly scale to the more expensive training of DL models. To achieve better anytime performance, one approach is to use multi-fidelity optimization, which approximates the full optimization problem by proxy tasks on cheaper fidelities, e.g. training only for a few epochs. Although there is evidence that correlation between different fidelities can be weak~\cite{ying-icml19a,yang-iclr20a}, we show in this paper that it is nevertheless possible to build a robust AutoDL system with strong anytime performance.

To study multi-fidelity approaches in a systematic way, we also introduce a new benchmark on learning curves in this paper, which we dub \emph{LCBench}. This new benchmark allows the community and us to study in an efficient way the challenges and the potential for using multi-fidelity approaches with DL and combining it with meta-learning, by providing learning curves of $2\,000$ configurations on a joint optimization space for each of $35$ diverse datasets.

Finally, we introduce \emph{Auto-PyTorch Tabular}, an AutoML framework targeted at tabular data that performs multi-fidelity optimization on a joint search space of architectural parameters and training hyperparameters for neural nets. Auto-PyTorch Tabular, the successor of Auto-Net~\cite{mendoza-automl16a} (part of the winning system in the first AutoML challenge~\cite{guyon-ijcnn15a}), combines state-of-the-art approaches from multi-fidelity optimization~\cite{falkner-icml18a}, ensemble learning~\cite{feurer-aaai15a} and meta-learning for a data-driven selection of initial configurations for warmstarting Bayesian optimization~\cite{feurer-automl18a}. 

Specifically the contributions of this paper are:

\begin{enumerate}
    \item We propose to combine BOHB, a robust optimizer for AutoDL, with automatically-designed portfolios of architectures \& hyperparameters and with ensembling, and show strong performance of this combination with Auto-PyTorch Tabular.
    \item We introduce \emph{LCBench}\footnote{\url{https://github.com/automl/LCBench}}, a new benchmark for studying multi-fidelity optimization w.r.t. learning curves on a joint optimization space of architectural and training hyperparameters across 35 datasets. In particular, we study the correlation between fidelities and evaluate hyperparameter importances across different fidelities to inform the design of an efficient AutoDL framework.
    \item To achieve state-of-the-art performance on tabular data, Auto-PyTorch Tabular combines tuned DNNs with simple traditional ML baselines.
    \item We introduce our \emph{Auto-PyTorch Tabular}\footnote{\url{https://github.com/automl/Auto-PyTorch}} tool to make deep learning easily applicable to new tabular datasets for the DL framework PyTorch. We show that Auto-PyTorch Tabular performs as well or even better than several other common AutoML frameworks: AutoKeras, AutoGluon, auto-sklearn and hyperopt-sklearn.
\end{enumerate}

In contrast to the mainstream in DL, here we do not focus on raw (e.g., image) data, but on tabular data, since we feel that DL is understudied for many applications of great relevance that feature tabular data, such as climate science,  medicine, manufacturing, finance, recommender systems, HR management, etc. While we focus the final AutoML system we propose on such tabular data, we also demonstrate the generality of our underlying methods on the well-known object recognition neural architecture search benchmark NAS-Bench-201~\cite{dong-iclr20a}.
\section{Related Work}\label{sec:problem_statement}

In the early years of automated machine learning, AutoML tools such as Auto-WEKA~\cite{thornton-kdd13a}, hyperopt-sklearn~\cite{komer-automl14a},
auto-sklearn~\cite{feurer-aaai15a} and
TPOT~\cite{olson-gecco16a}, focused on traditional machine learning pipelines because tabular data was and still is important for major applications. Popular choices for the optimization approach under the hood include Bayesian optimization, evolutionary strategies and reinforcement learning~\cite{automl-book-hpo}. In contrast to plain hyperparameter optimization, these tools can also address the combined algorithm selection and hyperparameter (CASH) problem by choosing several algorithms of an ML pipeline and their corresponding hyperparameters~\cite{thornton-kdd13a}. Although we focus on single-level hyperparameter optimization, our approach also addresses mixed configuration spaces with hierarchical conditional structures, e.g., the choice of the optimizer as a top-level hyperparameter and its sub-hyperparameters.

Another trend in AutoML is neural architecture search~\cite{elsken-jmlr19a}, which addresses the problem of determining a well-performing architecture of a DNN for a given dataset. 
Early NAS approaches established new state-of-the-art performances~\cite{stanley2002evolving, zoph-iclr17a}, but were also very expensive in terms of GPU compute resources. To improve efficiency, state-of-the-art systems make use of cell-search spaces~\cite{pham-icml18a}, i.e., configuring only repeated cell-architectures instead of the global architecture, and use gradient based optimization~\cite{liu-iclr19}. Although gradient-based approaches are very efficient on some datasets, they are known for their sensitivity to hyperparameter settings and instability~\cite{zela-iclr20a}. Recent practices for doing hyperparameter optimization only as a post-processing step of NAS ignore interaction effects between hyperparameter settings and choice of architecture and also are an expensive additional step at the end.

AutoNet~\cite{mendoza-automl16a}, based on Lasagne and Theano, was one of the first approaches jointly optimizing architectures and hyperparameters at scale and showed its success in the first AutoML Challenge. The intermediate step from AutoNet to a more robust and well-performing general AutoDL system was AutoNet~2.0, so far only described in a book chapter~\cite{mendoza2019towards}. Here, we formalize the vague description in that work and substantially improve it further to yield Auto-PyTorch, which combines the multi-fidelity optimization and ensembling approaches already present in Auto-Net~2.0 with meta-learning for warmstarting and an efficiently designed configuration space, where we make use of repeated blocks and groups and minimize the number of hyperparameters by only describing their shapes. As our experiments show (see Table \ref{tab:comparison}), this allows Auto-PyTorch to achieve much better performance than Auto-Net~2.0.

There is also a growing interest in well-designed AutoML benchmarks to take reproducibility and comparability of AutoML approaches into account~\cite{lindauer-arxiv19a}. For example, HPOlib~\cite{eggensperger-bayesopt13} provides benchmarks for hyperparameter optimization, ASlib~\cite{bischl-aij16a} for meta-learning of algorithm selection and NASBench-101~\cite{ying-icml19a}, NASBench-1Shot1~\cite{zela-iclr20b}, and NASBench-201~\cite{dong-iclr20a} for neural architecture search. However, to the best of our knowledge, there is yet no multi-fidelity benchmark on learning curves for the joint optimization of architectures and hyperparameters. We address this problem by our new LCBench, which provides data on 35 datasets in order to allow combinations with meta-learning.

Last but not least, several papers address the problem of understanding the characteristics of AutoDL tasks~\cite{sharma-dis19a,yang-iclr20a}.
A typical finding, for example, is that the design space is over-engineered, leading to very simple optimization tasks where even random search can perform well. As we show in our experiments, this does not apply to our tasks.
In addition to studying the characteristics of LCBench and also the full design space of Auto-PyTorch Tabular, we study hyperparameter importance from a global~\cite{rijn-kdd18a,hutter-icml14a,sharma-dis19a}, but also from a local point of view~\cite{biedenkapp-lion18a}, showing that both provide different insights.
\section{Auto-PyTorch}\label{sec:autopytorch}

We now present Auto-PyTorch, an AutoDL framework utilizing multi-fidelity optimization to jointly optimize NN architectural parameters and training hyperparameters. As the name suggests, Auto-PyTorch relies on the PyTorch framework~\cite{pytorch} as the DL framework. Auto-PyTorch implements and automatically tunes the full DL pipeline, including data preprocessing, neural architecture, network training techniques and regularization methods. Furthermore, it offers warmstarting the optimization by sampling configurations from portfolios as well as an automated ensemble selection. In the following, we provide more background on the individual components of our framework and introduce two search spaces of particular importance for this work. In this paper, we focus on tabular datasets in Auto-PyTorch Tabular, and while our methods are general, the configuration spaces we discuss in the following are geared to this case of tabular data. 

\subsection{Configuration Space}
\label{subsec:configspace}

The configuration space provided by Auto-PyTorch Tabular contains a large number of hyperparameters, ranging from preprocessing options (e.g. encoding, imputation) architectural hyperparameters (e.g. network type, number of layers) to training hyperparameters (e.g. learning rate, weight decay).
Specifically, the configuration space $\Lambda$ of Auto-PyTorch consists of all these hyperparameters and is structured by a conditional hierarchy, such that top-level hyperparameters can activate or deactivate sub-level hyperparameters based on their settings. Auto-PyTorch utilizes the ConfigSpace~\cite{lindauer2019a} package to represent its configuration space.

We propose to study two configuration spaces: (i) a small space that contains only few important design decisions, allowing for a thorough, fast and memory-efficient study; (ii) the full space of Auto-PyTorch Tabular which allows to achieve state-of-the-art performance.

\subsubsection{Search Space 1}
\label{sssec:space1}
For our smaller configuration space, we consider 7 hyperparameters in total. See Table~\ref{tab:space1}. Out of these, 5 are common training hyperparameters when training with momentum SGD. For the architecture, we deploy shaped MLPs~\cite{kotila-autonomio} with ReLUs and dropout~\cite{srivastava-jmlr14a}. Shaped MLP nets offer an efficient design space by avoiding layer-wise hyperparameters. We use a funnel shaped variant that only requires a predefined number of layers and a maximum number of units as input. The first layer is initialized with the maximum amount of neurons and each subsequent layer contains 

\begin{equation}
  n_{i} = n_{i-1} - \frac{\left( n_{max} -
  n_{out} \right)}{\left( n_{layers}-1 \right)}
\end{equation}

neurons, where $n_{i-1}$ is the number of neurons in the previous layer and $n_{out}$ is the number of classes. An example is depicted in Figure~\ref{fig:shapednets}.

\begin{table}[tbhp]
\caption{Hyperparameters and ranges of our Configuration Space 1.}
\centering
\begin{tabular}{@{}l|lccc@{}}
\toprule
                                                                                        & Name                 & Range            & log                 & Type  \\ \midrule
\multirow{3}{*}{\begin{tabular}[c]{@{}l@{}}Architecture\end{tabular}}  & number of layers     & {[}1, 5{]}       & -                         & int   \\
                                                                                        & max. number of units & {[}64, 512{]}    & \checkmark & int   \\
 \midrule
\multirow{4}{*}{\begin{tabular}[c]{@{}l@{}}Hyper-\\ parameters\end{tabular}} & batch size           & {[}16, 512{]}    & \checkmark & int   \\
                                                                                        & learning rate (SGD)  & {[}1e-4, 1e-1{]} & \checkmark & float \\
                                                                                        & L2 regularization   & {[}1e-5, 1e-1{]} & -                         & float \\
                                                                                        & momentum             & {[}0.1, 0.99{]}  & -                         & float \\
                                         & max. dropout rate    & {[}0,1.0{]}      & -                         & float \\ 
                                         \bottomrule
\end{tabular}
\label{tab:space1}
\end{table}

\subsubsection{Search Space 2}
\label{sss:space2}

\begin{figure}[tbh]
    \caption{Illustrating Shaped MLPs (left) and Shaped ResNets (right). The Shaped MLP is built with 100 max. units and 4 layers. The Shaped ResNet is built with 2 blocks, 1 block per group, and 100 max. units.}
    \centering
    \includegraphics[width=0.48\textwidth]{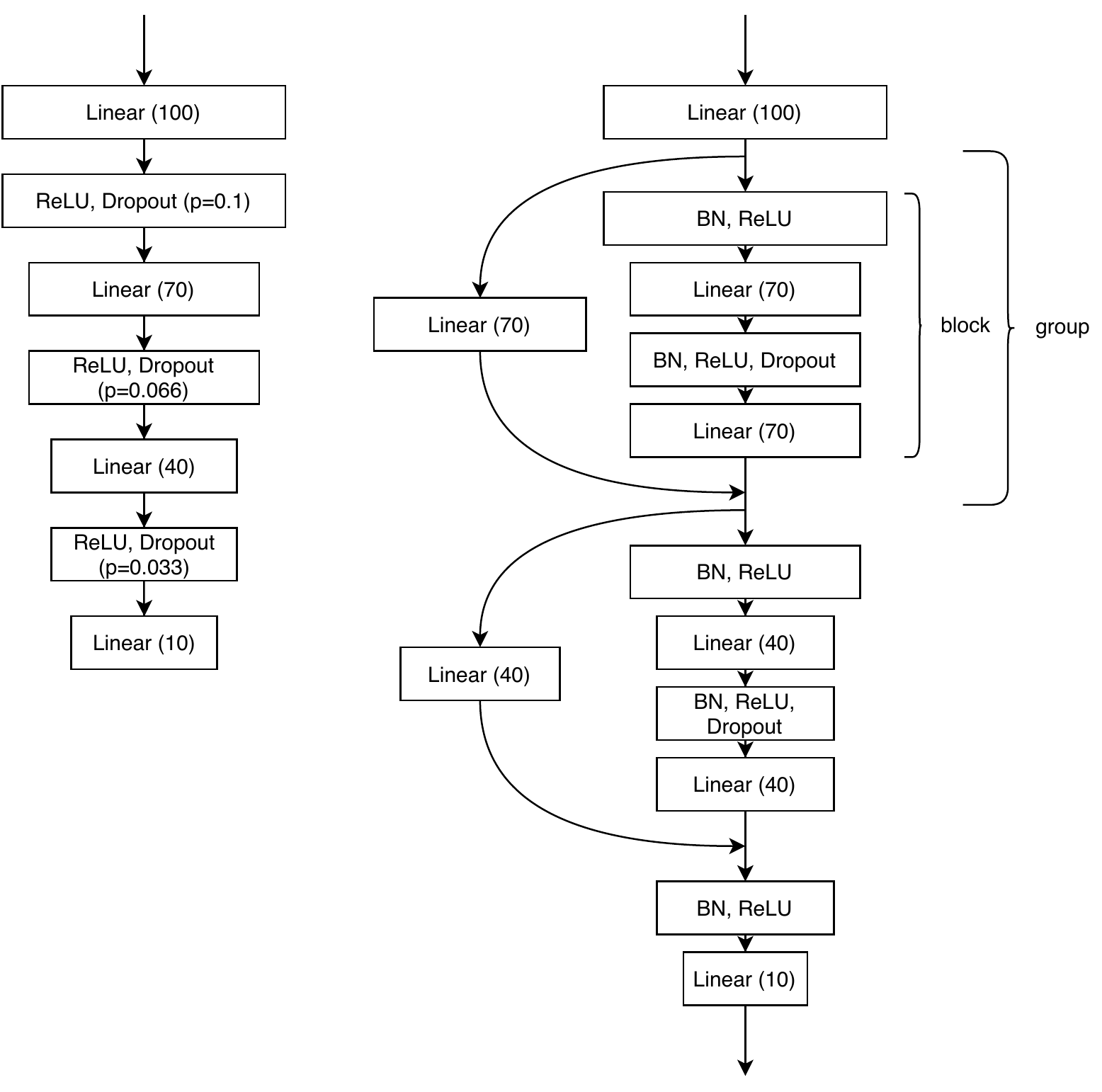}
    \label{fig:shapednets}
\end{figure}

To design a search space capable of finding high performance solutions, we extend the search space to include another training algorithm, Adam~\cite{kingma-iclr15a} and include mixup~\cite{zhang2017-mixup}, Shake-Shake~\cite{gastaldi2017-shakeshake} and ShakeDrop~\cite{yamada2018-shakedrop} to increase the choice of training techniques. We add preprocessing options with truncated SVD to deal with sparse datasets and add funnel-shaped residual networks (ResNets)~\cite{he2015-resnet}, allowing to train deeper networks. Like shaped MLPs, shaped ResNets allow for an efficient hyperparametrization by repeating groups with a predefined number of ResNet blocks. The output dimension of each group is determined identically to a shaped MLP with the equivalent number of layers. An example is provided in Figure~\ref{fig:shapednets}. All parameters of the larger search space are listed in Table~\ref{tab:large_cs}.

\begin{table}[tbh]
\caption{Hyperparameters and ranges of our Configuration Space 2}
\begin{tabular}{@{}l|lcccc@{}}
\toprule
                                                                                                & Name                   & Range                                                             & log       & type  & cond.     \\ \midrule
                                                                                                & network type           & \begin{tabular}[c]{@{}c@{}}{[}ResNet,\\  MLPNet{]}\end{tabular}   & -         & cat   & -         \\
                                                                                                & num layers (MLP)       & {[}1, 6{]}                                                        & -         & int   & \checkmark \\
                                                                                                & max units (MLP)        & {[}64, 1024{]}                                                    & \checkmark & int   & \checkmark \\
\multirow{5}{*}{\begin{tabular}[c]{@{}l@{}}Archi- \\tecture\end{tabular}}  & max dropout (MLP)      & {[}0, 1{]}                                                        & -         & float & \checkmark \\
                                                                                                & num groups (Res)       & {[}1, 5{]}                                                        & -         & int   & \checkmark \\
                                                                                                & blocks per group (Res) & {[}1, 3{]}                                                        & -         & int   & \checkmark \\
                                                                                                & max units (Res)        & {[}32, 512{]}                                                     & \checkmark & int   & \checkmark \\
                                                                                                & use dropout (Res)      & {[}F, T{]}                                                        & -         & bool  & \checkmark \\
                                                                                                & use shake drop         & {[}F, T{]}                                                        & -         & bool  & \checkmark \\
                                                                                                & use shake shake        & {[}F, T{]}                                                        & -         & bool  & \checkmark \\
                                                                                                & max dropout (Res)      & {[}0, 1{]}                                                        & -         & float & \checkmark \\
                                                                                                & max shake drop (Res)   & {[}0, 1{]}                                                        & -         & float & \checkmark \\ \midrule
                                                                                                & batch size             & {[}16, 512{]}                                                     & \checkmark & int   & -         \\
                                                                                                & optimizer              & {[}SGD, Adam{]}                                                   & -         & cat   & -         \\
                                                                                                & learning rate (SGD)    & {[}1e-4, 1e-1{]}                                                  & \checkmark & float & \checkmark \\
\multirow{5}{*}{\begin{tabular}[c]{@{}l@{}}Hyper-\\ para-\\ meters\end{tabular}} & L2 reg. (SGD)     & {[}1e-5, 1e-1{]}                                                  & -         & float & \checkmark \\
                                                                                                & momentum               & {[}0.1, 0.999{]}                                                  & -         & float & \checkmark \\
                                                                                                & learning rate (Adam)   & {[}1e-4, 1e-1{]}                                                  & \checkmark & float & \checkmark \\
                                                                                                & L2 reg. (Adam)    & {[}1e-5, 1e-1{]}                                                  & -         & float & \checkmark \\
                                                                                                & training technique     & \begin{tabular}[c]{@{}c@{}}{[}standard,\\  mixup{]}\end{tabular}  & -         & cat   & -         \\
                                                                                                & mixup alpha            & {[}0, 1{]}                                                        & -         & float & \checkmark \\
                                                                                                & preprocessor           & \begin{tabular}[c]{@{}c@{}}{[}none,\\  trunc. SVD{]}\end{tabular} & -         & cat   & -         \\
                                                                                                & SVD target dim         & {[}10, 256{]}                                                     & -         & int   & \checkmark\\
                            \bottomrule
\end{tabular}
\label{tab:large_cs}
\end{table}

\subsection{Multi-Fidelity Optimization}
\label{subsec:bohb}
Since training DNNs can be expensive, achieving good anytime performance is a challenging task. Auto-PyTorch follows a multi-fidelity optimization approach to tackle this, using BOHB~\cite{falkner-icml18a} to find well-performing configurations over multiple budgets. BOHB combines Bayesian optimization (BO)~\cite{mockus-jgo94} with Hyperband (HB)~\cite{li-jmlr18a} and has been shown to outperform BO and HB on many tasks. It also achieves speed ups of up to 55x over Random Search~\cite{falkner-icml18a}.

Similar to HB, BOHB consists of an outer loop and an inner loop. In the inner loop, configurations are evaluated first on the lowest budget (defined by the current outer loop step) and well-performing configurations advance to higher budgets via SuccessiveHalving (SH)~\cite{jamieson-aistats16}. In the outer loop, the budgets are calculated from a given minimum budget $b_{min}$, a maximum budget $b_{max}$ and a scaling factor $\eta$ such that for two successive budgets $\frac{b_{i+1}}{b_{i}} = \eta$.

\begin{figure}[htp]
    \centering
    \caption{Components and Pipeline of a parallel Auto-PyTorch based on a master-work principle.}
    \includegraphics[width=0.35\textwidth]{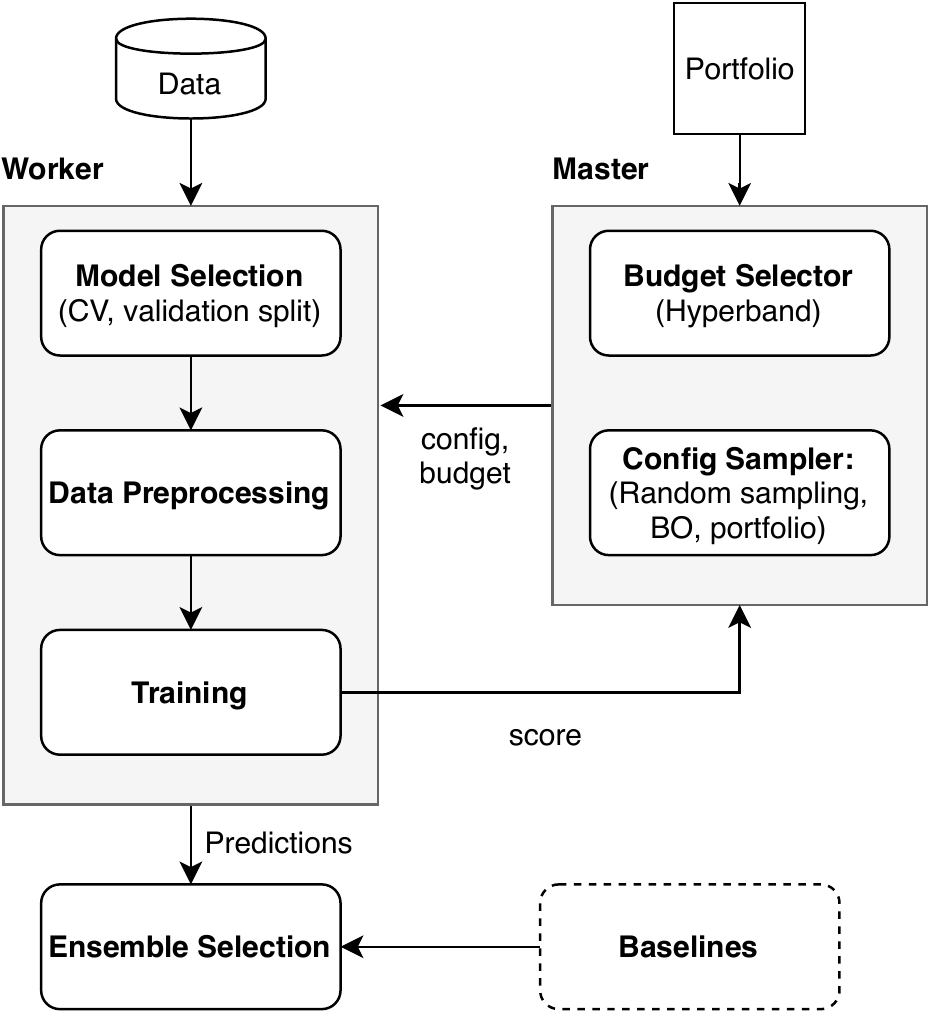}
    \label{fig:apt}
\end{figure}

Unlike HB, BOHB fits a kernel density estimator (KDE) as a probabilistic model to the observed performance data and uses this to identify promising areas in the configuration space by trading off exploration and exploitation. Additionally, a fraction of configurations is sampled at random to ensure convergence. Before fitting a KDE on budget $b$, BOHB samples randomly until it has evaluated as many data points on $b$ as there are hyperparameters in $\Lambda$ to ensure that the BO model is well-initialized. For sampling a configuration from the KDE, BOHB uses the KDE on the highest available budget. 
Since the configurations on smaller budgets are evaluated first (according to SH), the KDEs on the smaller budgets are available earlier and can guide the search until better informed models on higher budgets are available.

An important choice in setting up multi-fidelity optimization is the type of budget used to create proxy tasks. There are many possible choices such as runtime, number of training epochs, dataset subsamples~\cite{klein-aistats17} or network architecture-related choices like number of stacks~\cite{liu-iclr19}. We choose the number of training epochs as our budgets over the runtime because of its generality and interpretability. Epochs allow a simple comparison between performance of configurations across datasets and transfer of configurations (e.g. learning rate schedules).

\subsection{Parallel Optimization}
Since BOHB samples from its KDE instead of meticulously optimizing the underlying acquisition function, a batch of sampled configurations has some diversity. As shown by Falkner et al.~\cite{falkner-icml18a}, this approach leads to efficient scaling for a parallel optimization. We make use of this by deploying a master-worker architecture, such that Auto-PyTorch can efficiently use additional compute resources; see Figure~\ref{fig:apt}.

\subsection{Model Selection}
To ensure model generalization, a good model selection strategy is required in any AutoML framework. To that end, Auto-PyTorch implements a variety of options. It supports the hold-out protocol, allowing for input of user-defined splits or automated splitting, or cross-validation with any of scikit-learn's~\cite{scikit-learn} cross-validation iterators. The model performance is then evaluated via a predefined, or user-specified metric, on the validation set and used to select well-performing configurations. Auto-PyTorch also offers using early stopping via the performance on the validation set to further improve generalization.

\subsection{Ensembles} \label{subsec:ensemles}
The iterative nature of Auto-PyTorch implies that many models are trained and evaluated. This naturally allows for further boosting the predictive performance by ensembling the evaluated models. Auto-PyTorch uses an approach for ensembling inspired by auto-sklearn~\cite{feurer-nips2015a}, implementing an automated post-hoc ensemble selection method~\cite{caruana-icml04a} which efficiently operates on model predictions stored during optimization. Starting from an empty set, the ensemble selection iteratively adds the model that gives the largest performance improvement until reaching a predefined ensemble size. Allowing multiple additions of the same model results in a weighted ensemble. Since it has been shown that regularization improves ensemble performance~\cite{caruana-icdm06a}, and following auto-sklearn~\cite{feurer-nips2015a} we only consider the $k$ best models in our ensemble selection ($k=30$ here). 

An advantage of our post-hoc ensembling is that we can also include other models besides DNNs easily. As we show in our experiments, this combination of different model classes is one of the key factors to achieve state-of-the-art performance on tabular data.

\subsection{Portfolios}
\label{subsec:portfolios}

BOHB is geared towards good anytime performance on large search spaces. However, it starts from scratch for each new task. 
Therefore, we warmstart the optimization to improve the early performance even further. Auto-sklearn used task meta-features to determine promising configurations for warmstarting~\cite{feurer-aaai15a}. In this work, we follow the approach of the newer PoSH-Auto-Sklearn to utilize \emph{portfolios} in a meta-feature free approach. Auto-PyTorch simply starts BOHB's first iteration with a set of complementary configurations that cover a set of meta-training datasets well; afterwards it transitions to BOHB's conventional sampling.

To construct the portfolio $\mathcal{P}$ offline, we performed a BOHB run on each of many meta training datasets $\mathbf{D}_{\text{meta}}$, giving rise to a set of portfolio candidates $\mathcal{C}$. The incumbent configurations from the individual runs are then evaluated on all $\mathcal{D}\in \mathbf{D}_{\text{meta}}$, resulting in a performance meta-matrix, such that the portfolio can simply be constructed by table look-ups. From the candidates $\lambda \in \mathcal{C}$, configurations $\lambda$ are iteratively and greedily added to the portfolio $\mathcal{P}$ in order to minimize its mean relative regret $\mathcal{R}$ over all meta datasets $\mathbf{D}_{\text{meta}}$~\cite{xu-aaai10a,xu-rcra11a}. Specifically, we successively add $\lambda_i^*$ to the previous portfolio $\mathcal{P}_{i-1}$ in iteration $i$:

\begin{equation*}
    \lambda_i^* \in \mathop{\mathrm{arg\,min}}_{\lambda \in \mathcal{C}}
    \sum_{\substack{\left(\mathcal{D}_{train},
    \mathcal{D}_{test}\right) \\ \in \mathbf{D}_{\text{meta}}}} 
    \min_{\lambda' \in \mathcal{P}_{i-1} \bigcup \{\lambda \}}
    \mathcal{R}\left( \lambda', \mathcal{D}_{train}, \mathcal{D}_{test} \right)
\end{equation*}

\noindent
where $\mathcal{P}_0$ is the empty portfolio and the relative regret $\mathcal{R}$ is calculated w.r.t. the best observed performance over the portfolio candidates. Therefore, in the first iteration the configuration that is best on average across all datasets is added to the portfolio. In all subsequent iterations, configurations are added that tend to be more specialized to subsets of $\mathbf{D}_{\text{meta}}$ for which further improvements are possible.

Configurations are added in this manner until a predefined portfolio size is reached. Limiting the size of the portfolio in this manner balances between warmstarting with promising configurations and the overhead induced by first running the portfolio.

This approach assumes (as all meta-learning approaches) that we have access to a reasonable set of meta-training datasets that are representative of meta-test datasets. We believe that this is particularly possible for tabular datasets because of platforms such as OpenML~\cite{vanschoren-sigkdd14a}, but sizeable dataset collections also already exist for several other data modalities, such as images and speech.
\section{LCBench: Comprehensive AutoDL Study for Multi-Fidelity Optimization}\label{sec:exploration}

To gain insights on how to design multi-fidelity optimization for AutoDL, we first performed an initial analysis on the smaller configuration space (see Section~\ref{sssec:space1}). Starting on a small configuration space allows us to sample configurations more densely and with many repeats to obtain information about all regions of the space. We study (i) the performance distribution of configurations across datasets to show that single configurations can perform well across datasets, (ii) the correlation between budgets for adaptive and non-adaptive learning rate schedules to justify the use of multi-fidelity optimization and (iii) the importance of architectural design decisions and  hyperparameters across budgets. Based on our findings we suggest a number of design choices for the full Auto-PyTorch Tabular design space. In particular, we argue for the benefit of portfolios and justify the use of multi-fidelity optimization.

\subsection{Experimental Setup}
\label{subsec:setup_lcbench}

We collected data by randomly sampling $2\,000$ configurations and evaluating each of them across 35 datasets and three budgets. Each evaluation is performed with three different seeds on Intel Xeon Gold 6242 CPUs with one core per evaluation, totalling $1\,500$ CPU hours.

\subsubsection{Datasets}
We evaluated the configurations on 35 out of 39 datasets from the AutoML Benchmark~\cite{gijsbers-automl} hosted on OpenML~\cite{vanschoren-sigkdd14a}. As we designed LCBench to be a cheap benchmark, we omit the four largest datasets (robert, guillermo, riccardo, dilbert), in order to ensure low runtimes and memory footprint. Nevertheless, the datasets we chose are very diverse in the number of features ($5-1\,637$), data points ($690-581\,012$) and classes ($2-355$) (also see Figure~\ref{fig:feat_vs_inst}) and cover binary and multi-class classification, as well as strongly imbalanced tasks. Whenever possible, we use the given test split  with a $33~\%$ test split and additionally use fixed $33~\%$ of the training data as validation split. In case there is no such OpenML task with a $33~\%$ split available for a dataset, we create a $33~\%$ test split and fix it across the configurations.

\subsubsection{Configuration Space and Training}
We searched on Search Space 1 introduced in Section~\ref{sssec:space1} and Table~\ref{tab:space1}, which comprises $7$~hyperparameters ($3$~integer, $4$~float), two of which describe the architecture and five of which are training hyperparameters. We used SGD for training and cosine annealing~\cite{loshchilov-iclr17a} as a learning rate scheduler. We fixed all remaining hyperparameters to their default.

\subsubsection{Budgets}
We used the number of epochs as our budgets and evaluated each configuration for $12$, $25$ and $50$ epochs. These budgets correspond to the ones BOHB selects with the parameters $\left( b_{\text{min}}, b_{\text{max}}, \eta \right) = \left(12, 50, 2\right)$. For each of these evaluations, the cosine annealing scheduler was set to anneal to $10^{-8}$ when reaching the respective budget. Since we also log the learning curve, we hence are able to compare the adaptive learning rate scheduling with a seperate cosine schedule for each budget with the non-adaptive one using a single cosine schedule for 50 epochs, also evaluated at 12 and 25 epochs.

\subsubsection{Data}
Throughout these evaluations we logged a vast number of metrics, including learning curves on performance metrics on train, validation and test set, configuration hyperparameters and global as well as layer wise gradient statistics over time like gradient mean and standard deviation. Although we only use the performance metrics and configurations in this study, the full data is publicly available and we hope that it will be very helpful for future studies by the community. We dub this benchmark LCBench.

\subsection{Results}

We study the following questions on LCBench in order to inform our design of Auto-PyTorch Tabular.

\begin{description}
    \item[\textbf{RQ1}] Are there configurations that perform well on several datasets?
    \item[\textbf{RQ2}] Is it possible to cover most datasets based on a few complementary configurations?
    \item[\textbf{RQ3}] Is there a strong correlation between the budgets if we use number of epochs as budgets?
    \item[\textbf{RQ4}] Is the importance of hyperparameters consistent across datasets such that models on different budgets can be efficiently used?
\end{description}

\begin{figure*}[!ht]
  \centering
  \includegraphics[width=0.48\linewidth]{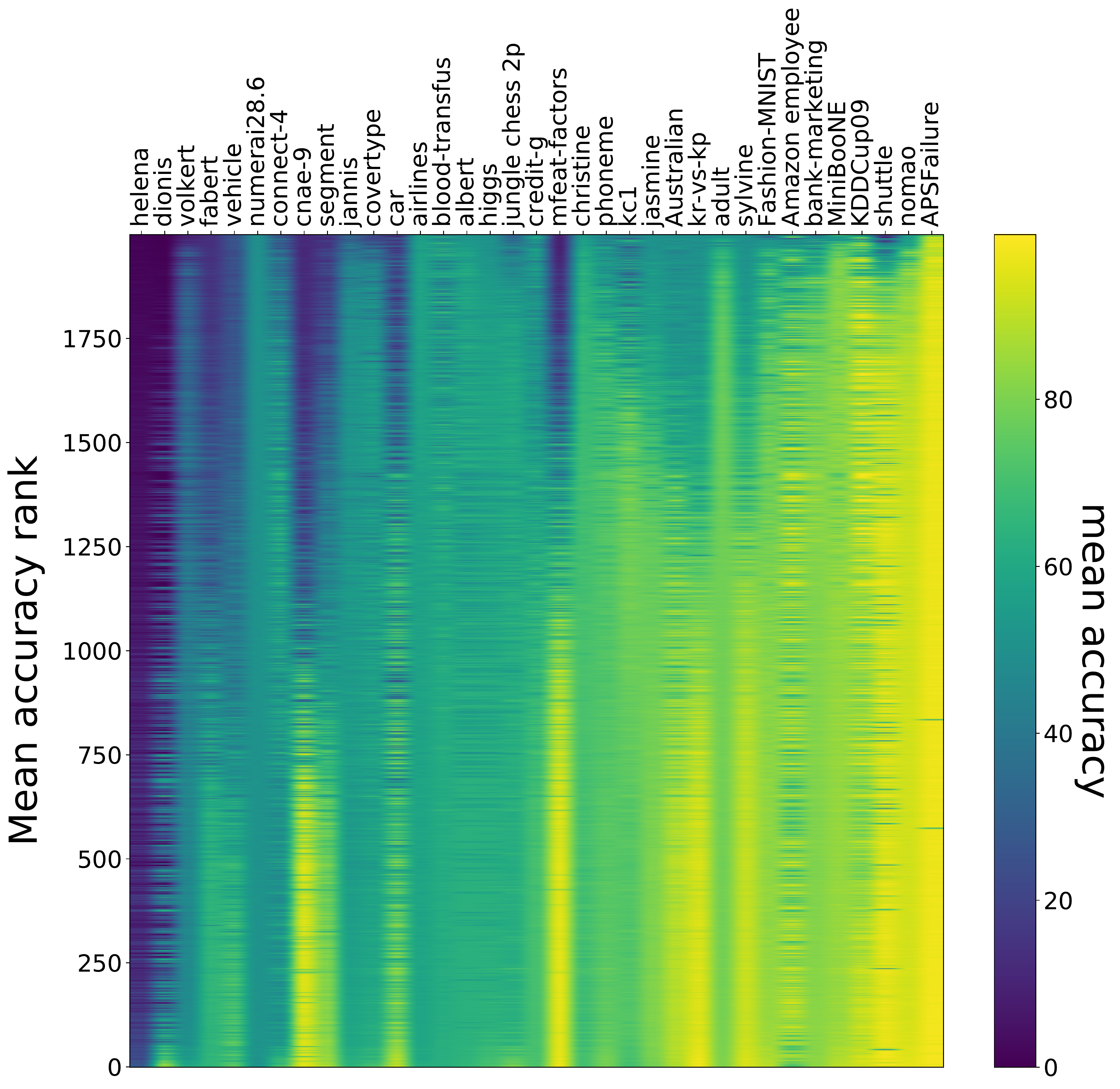}
  \includegraphics[width=0.48\linewidth]{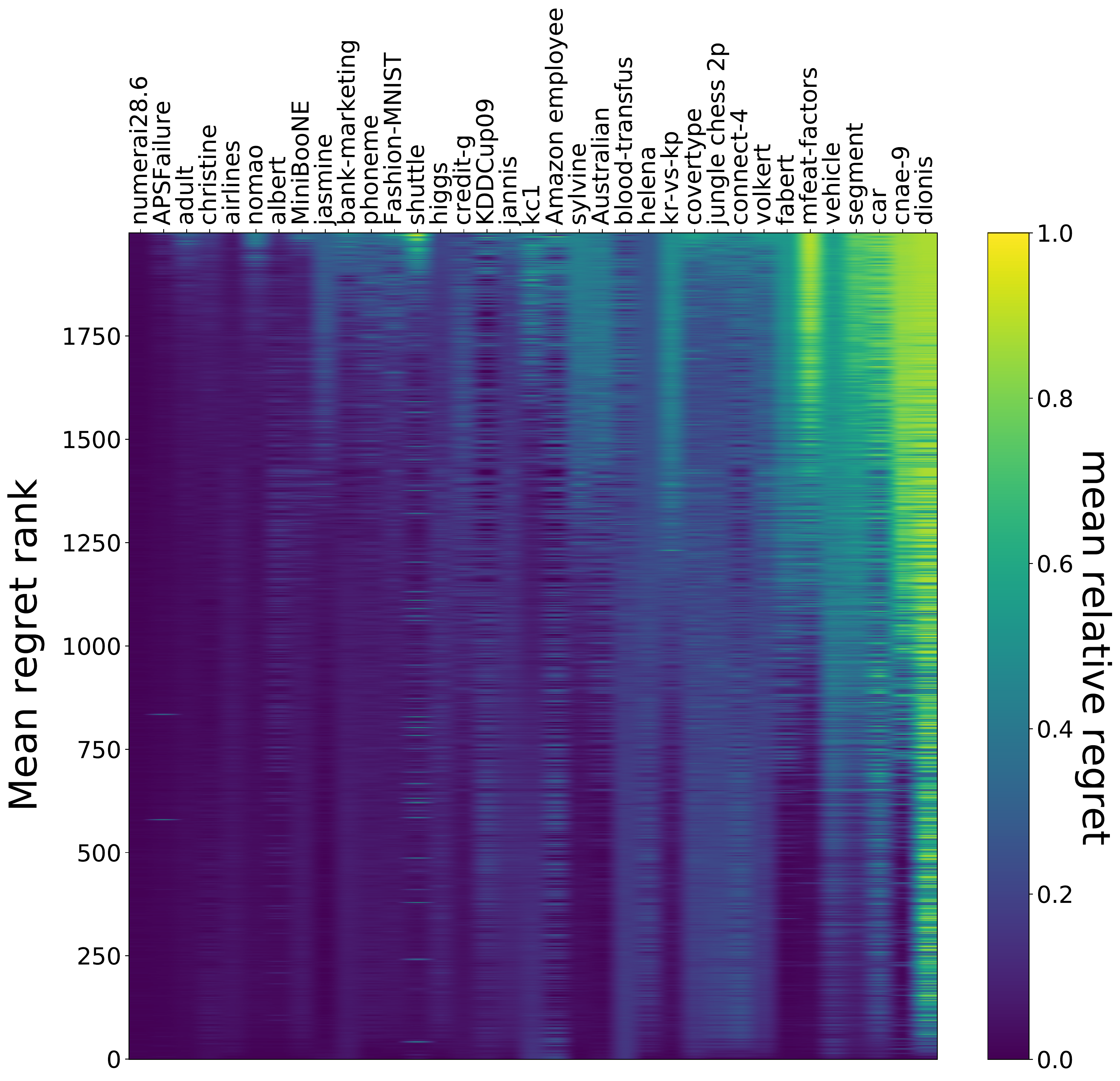}
  \caption{Mean validation accuracy (left) and mean relative regret (right) of 2000 evaluated configurations across 35 datasets. For better visualization, we sort by mean accuracy/regret along each axis respectively.}
\label{fig:lcb_full_matrix}
\end{figure*}

\subsubsection{RQ1: Performance Across Datasets}
Figure~\ref{fig:lcb_full_matrix} illustrates the performance of all 2000 configurations across all datasets. We show the mean accuracy on the validation set, as well as the mean relative regret with respect to the best observed validation accuracy, averaging over the 3 runs.
As can be seen, the performance distribution varies strongly across different tasks. On some, most configurations perform well (e.g. APSFailure) such that even a random sampling quickly yields good solutions. On the other hand, there are tasks where good configurations are sparse and random search is very inefficient (e.g. dionis).
Although better mean performance of a configuration does not strictly imply a better performance on each dataset, indicated by the stripe patterns in Figure~\ref{fig:lcb_full_matrix}, there clearly are configs which perform well across datasets. Always selecting the best configuration for each dataset would improve the average absolute error by $3\%$ compared to selecting the configuration that is best on average across all datasets. The standard deviation across the datasets is at $15.3\%$, indicating that some datasets benefit much more from selecting the best configurations than others.

Interestingly, selecting the best of the $2\,000$ evaluated configurations for each dataset shows that some of them are best for several datasets. Overall the final set of best configurations contains 22 unique configurations from 35 datasets with up to 7 occurrences of a single configuration.
This provides evidence that transferring  configurations to other datasets is very promising if the configuration is well selected. This is an argument for the use of well designed portfolios, as proposed in Section~\ref{subsec:portfolios}.

\subsubsection{RQ2: Performance of Portfolios with Different Sizes}

\begin{figure}[thbp]
    \centering
    \caption{Portfolio performance as a function of portfolio size on the $2\,000$ configurations from LCBench.}
    \includegraphics[width=0.42\textwidth]{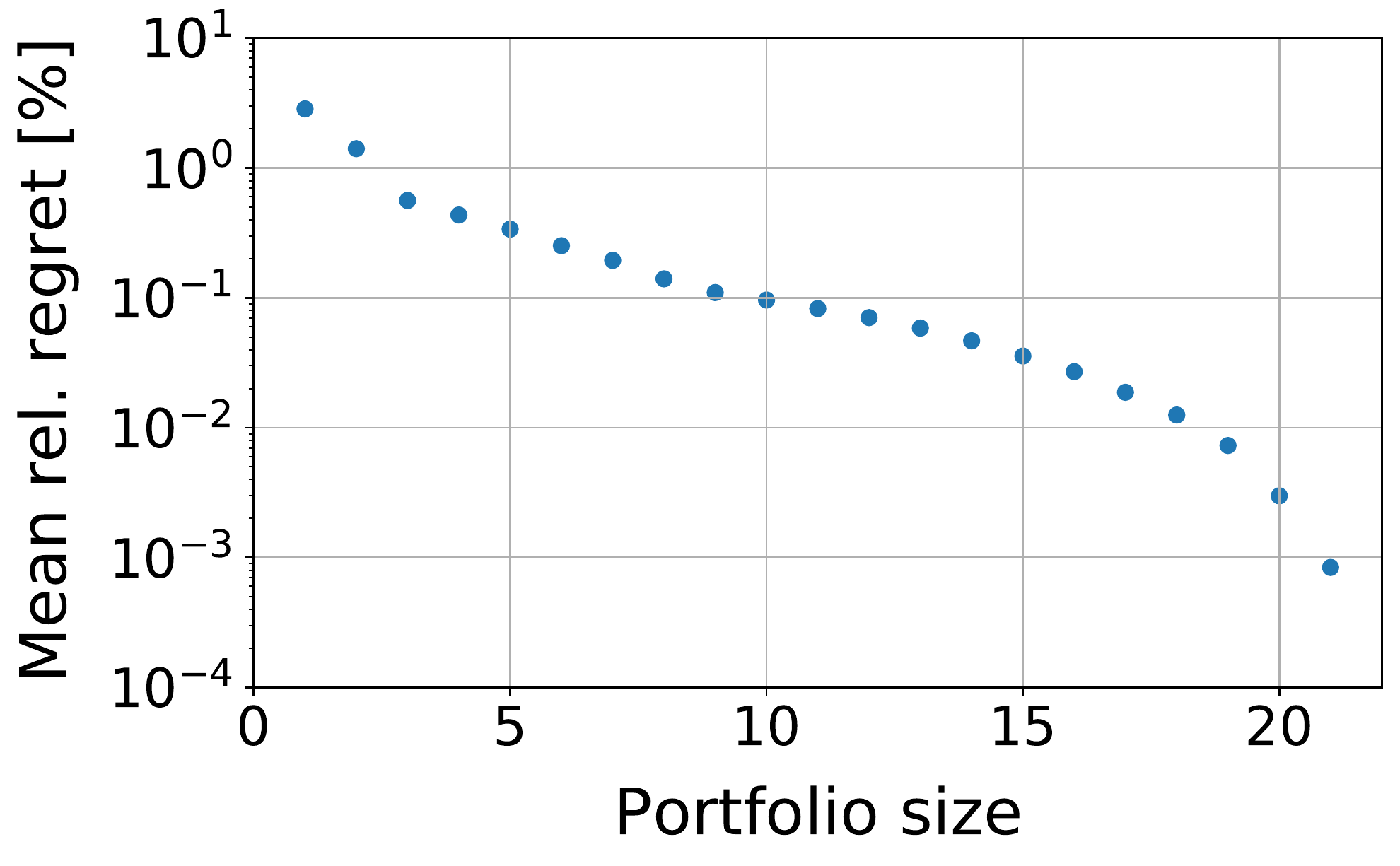}
    \label{fig:lcb_portfolio}
\end{figure}

The results of RQ1 already indicate that portfolios should be a simple but promising approach for achieving good performance with little effort. In Figure~\ref{fig:lcb_portfolio}, we show how the coverage of the portfolio (in terms of regret compared to the full portfolio of $2000$ configurations) depends on the size of the portfolio. Whereas a single configuration does not perform well on all datasets, a portfolio of size $10$ has an average accuracy regret of less than $0.1\%$ and thus covers all datasets to a sufficient degree.

\subsubsection{RQ3: Correlation Between Budgets}
\begin{figure}[thbp]
    \centering
    \caption{Average Spearman rank correlation for different datasets and budgets.}
    \includegraphics[width=0.48\textwidth]{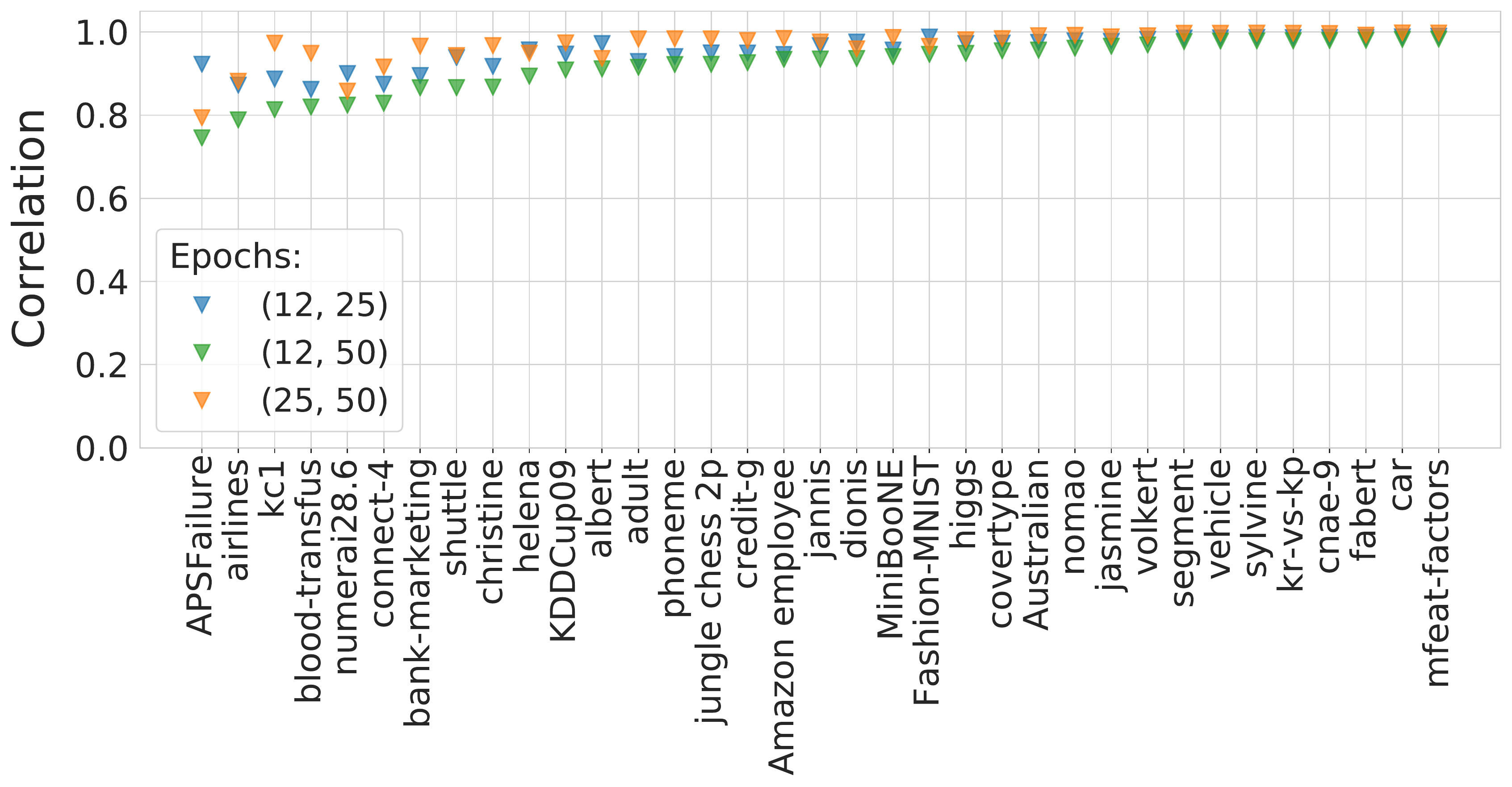}
    \label{fig:lcb_budget_corr}
\end{figure}

To enable multi-fidelity optimization, there has to be mutual information between different budgets. We thus study the correlation between budgets utilizing the Spearman rank correlation coefficient. More precisely, we choose a dataset $\mathcal{D} \in \mathbf{D}$, and compute $\tau$ for a budget pair $\left( b, b^{\prime} \right)$ over all configurations $\lambda \in \Lambda$ to obtain $\tau \left( \mathcal{D}, b, b^{\prime} \right)$. We perform this for all datasets and budget pairs.  Figure~\ref{fig:lcb_budget_corr} shows the corresponding results.
Generally, the correlations between budgets are rather high as required for multi-fidelity optimization. As expected, the adjacent budget pairs $\left( 12, 25 \right)$ and $\left( 25, 50 \right)$ exhibit a larger correlation than the more distant budget pair $\left( 12, 50 \right)$, see Table~\ref{tab:mean_corrs}.

Additionally, we consider learning rate schedules with and without adaptation to the budget at hand. For adaptive scheduling, we consider cosine annealing, where the epoch at which the minimum is reached is set to the current budget. We refer to the case where the minimum is always set to be reached at the maximum budget as non-adaptive scheduling. We observe that adaptive learning rate schedules exhibit a slightly worse correlation compared to non-adaptive strategies on the larger budgets.

\begin{table}[tpb]
\caption{Average Spearman rank correlation across datasets and configurations.}
\centering
\begin{tabular}{@{}l|ccc@{}}
\toprule
\begin{tabular}[c]{@{}l@{}}Budget pair\\ (epochs)\end{tabular} & \begin{tabular}[c]{@{}c@{}}Non-adaptive\\ scheduling\end{tabular} & \begin{tabular}[c]{@{}c@{}}Adaptive\\ scheduling\end{tabular} & \begin{tabular}[c]{@{}c@{}}Adaptive vs\\ Non-adaptive\end{tabular} \\ \midrule
(12, 25)                                                       & 0.94 $\pm$ 0.03                                                      & \textbf{0.95 $\pm$ 0.05}                                         & 0.89                                                               \\
(25, 50)                                                       & \textbf{0.96 $\pm$ 0.06}                                             & 0.94 $\pm$ 0.05                                                  & 0.86                                                               \\
(12, 50)                                                       & \textbf{0.91 $\pm$ 0.07}                                             & 0.88 $\pm$ 0.09                                                  & 0.71                                                               \\ \bottomrule
\end{tabular}
\label{tab:mean_corrs}
\end{table}

\subsubsection{RQ4: Hyperparameter Importance}

The analysis of hyperparameter importance provides insights into which hyperparameters are crucial  to achieve optimal performance. This is of particular interest (i) when jointly searching for the architecture and hyperparameters and (ii) when using multi-fidelity optimization to prevent the importance changing substantially across budgets. 

\begin{figure}[htp]
    \centering
    \caption{Boxplots on the hyperparameter importance according to fANOVA (left) and LPI (right) on LCBench.}
    \includegraphics[width=0.48\textwidth]{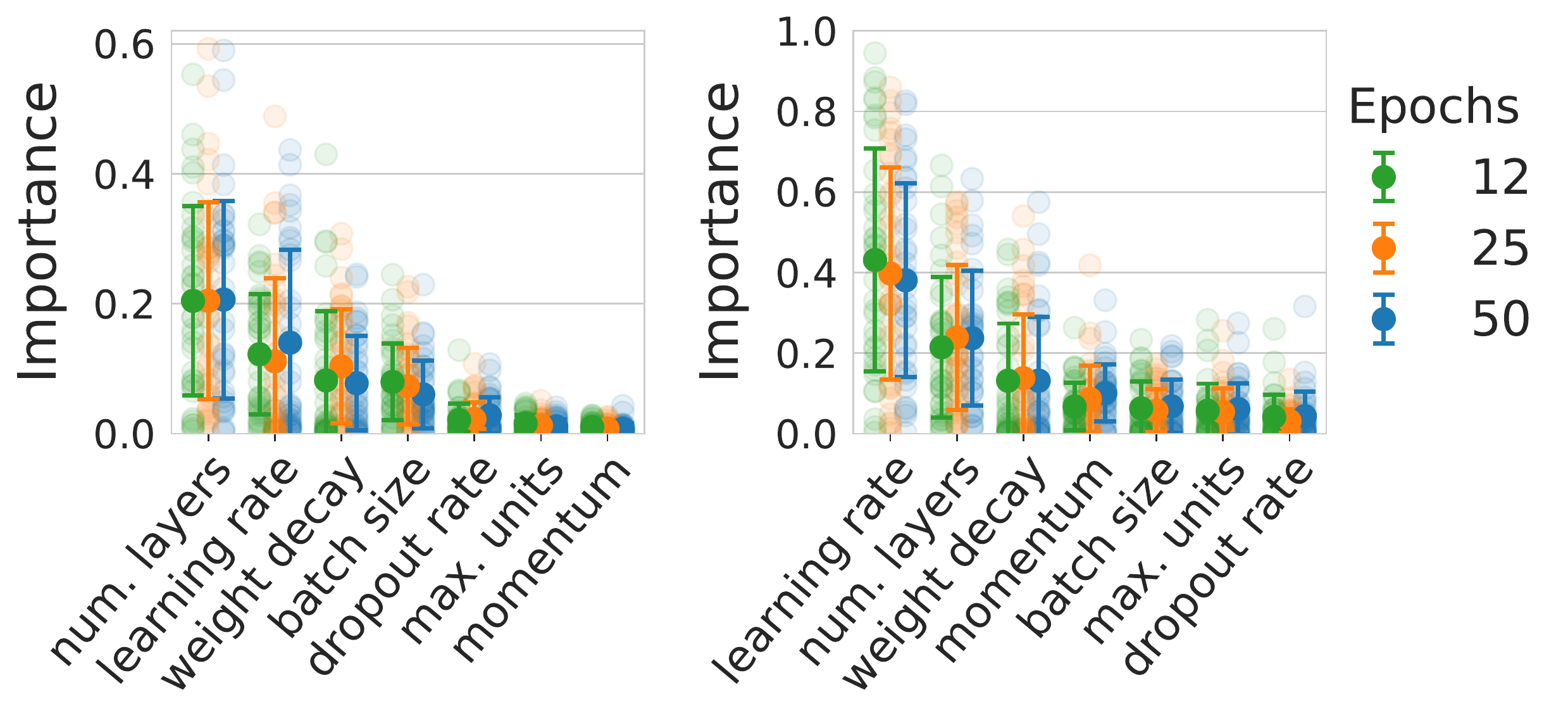}
    \label{fig:lcbench_pimp}
\end{figure}

We chose two approaches: a global analysis based on fANOVA~\cite{hutter-icml14a} and Local Hyperparameter Importance (LPI)~\cite{biedenkapp-lion18a}. 
Both approaches quantify importance as the variation caused by changing a single hyperparameter; fANOVA marginalizes over all other parameters in the configuration space, whereas LPI fixes all other values to a given (incumbent) configuration.
Both utilize a random forest as an empirical performance model~\cite{hutter-icml14a} fitted on the configurations and their observed performances. 
Generally, fANOVA is a global method, as it reasons about the entire configuration space by marginalizing out other effects.
Results of an fANOVA analysis can be seen in Figure~\ref{fig:lcbench_pimp}. We see that the design of the architecture and the hyperparameters significantly influences the performance of a configuration. Surprisingly, the number of layers (num layers) is the most important hyperparamter, even more important than learning rate or weight decay. However, the maximum number of neurons (max units) is less important, which can be explained by the fact that more layers also leads to overall more neurons by using our shaped networks; however more neurons (max. units) do not imply more layers.

As we expected, the importance of the learning rate slightly increases if we train for longer (i.e. larger budgets). The weight decay and batch size have a small trend to decrease in importance instead. All other importance scores are fairly constant across budgets. Overall, we can conclude that the hyperparameter importance according to fANOVA is quite stable across budgets.

LPI is inspired by the human approach in searching for performance improvements. It defines the importance of a hyperparameter $p$ as the variance caused by $p$ at an incumbent configuration normalized by the sum of the variances over all individual hyperparameters at that configuration.  In considering only the neighbourhood of one configuration, LPI marks an extremely local method along one axis.

In contrast to fANOVA, the LPI assigns the largest overall importance to the learning rate, see Figure~\ref{fig:lcbench_pimp}. This means that for a fixed incumbent network changing the learning rate will lead to more catastrophic performance losses, compared to changing e.g., the number of layers. While the number of layers and weight decay setting are ranked highly, the importance is more distributed overall. These importance scores again are quite stable across budgets.

Since the number of layers is of major importance, the addition of ResNets to the configuration space of Auto-PyTorch might yield better solutions, as they allow training much deeper DNNs more easily.
\section{Results of Auto-PyTorch}\label{sec:optimization}

We now move to evaluate the full Auto-PyTorch Tabular system, which combines the full configuration space (Section~\ref{subsec:configspace}), multi-fidelity optimization (Section~\ref{subsec:bohb}),  ensembles (Section~\ref{subsec:ensemles}) and portfolio generation (Section~\ref{subsec:portfolios}). To study the improvement effects of each of these, we study them isolated in a thorough ablation study.
We also compare Auto-PyTorch Tabular to various other AutoML systems and verify that Auto-PyTorch's general methodological core of multi-fidelity meta-learning also yields strong results for object recognition in NAS-Bench-201.

\subsection{Experimental Setup}

\subsubsection{Construction of the Portfolio}
To collect candidate configurations for a portfolio, we ran Auto-PyTorch on $100$ meta datasets $\mathbf{D}_{\text{meta}}$ from OpenML~\cite{vanschoren-sigkdd14a}. For details on selecting these datasets we refer to Appendix~\ref{app:datasets}.

We used the same budgets as before and searched in $24$ hours for at most $300$ BOHB iterations. To prevent long runtimes on large datasets with poor performances, we also enabled early stopping. We ran on the same hardware as described in Section~\ref{subsec:setup_lcbench} with $3$ workers and $2$ cores per worker, totalling in about $900$ CPU hours. The configuration space is the large space described in Section~\ref{sss:space2}.

As described in Section~\ref{subsec:portfolios}, we obtained a performance matrix with $100 \times 100$ entries by evaluating the incumbent configurations of each individual run on all other datasets from which we create a \emph{greedy portfolio} of size $16$. To have another portfolio baseline,  we also include all $100$ incumbent configurations in a portfolio, dubbed \emph{simple portfolio}.

\subsubsection{Evaluation Setup of Auto-PyTorch}

To evaluate the benefit of our framework components, we ran Auto-PyTorch on test datasets in an ablating fashion. We use the same number of cores, workers and epoch budgets as for the portfolio construction, but run up to 72 hours or 300 BOHB iterations. When running plain BO, we use the TPE-like BO implementation based on KDEs in BOHB, as described in Section~\ref{subsec:bohb}. We choose $8$ test datasets from OpenML roughly covering the feature and class distribution of the meta datasets. See Figure~\ref{fig:feat_vs_inst}.

For comparison, we evaluate a multitude of baselines on the test datasets. Following Erickson et al.~\cite{erickson2020autogluon}, we choose LightGBM~\cite{ke2017-lightgbm}, CatBoost~\cite{dorogush2018-catboost}, Random Forests~\cite{breimann-mlj01a}, Extremely Randomized Trees~\cite{geurts_ml2006a} and $k$-nearest-neighbours, and use the same hyperparameter setting. For the latter three, we use the scikit-learn~\cite{scikit-learn} implementations. For training the baselines, $6$~CPU cores were allocated to match the resources of an Auto-PyTorch run. We later also consider ensembles built from Auto-PyTorch models and these baselines. Finally, we compare to other common AutoML frameworks. We report the error obtained from the accuracy on the test set for all experiments.

\subsection{Studying the Large Configuration Space}

We now extend the analysis previously presented on the smaller configuration space to the configuration space of Auto-PyTorch to check which of the previous hypotheses also hold here. We do this study on the meta-train datasets, on which we collected a lot of meta-data for the portfolio construction, leaving the test datasets for a final evaluation.

\begin{figure}[htp]
    \centering
    \caption{Correlation between budgets.}
    \includegraphics[width=0.45\textwidth]{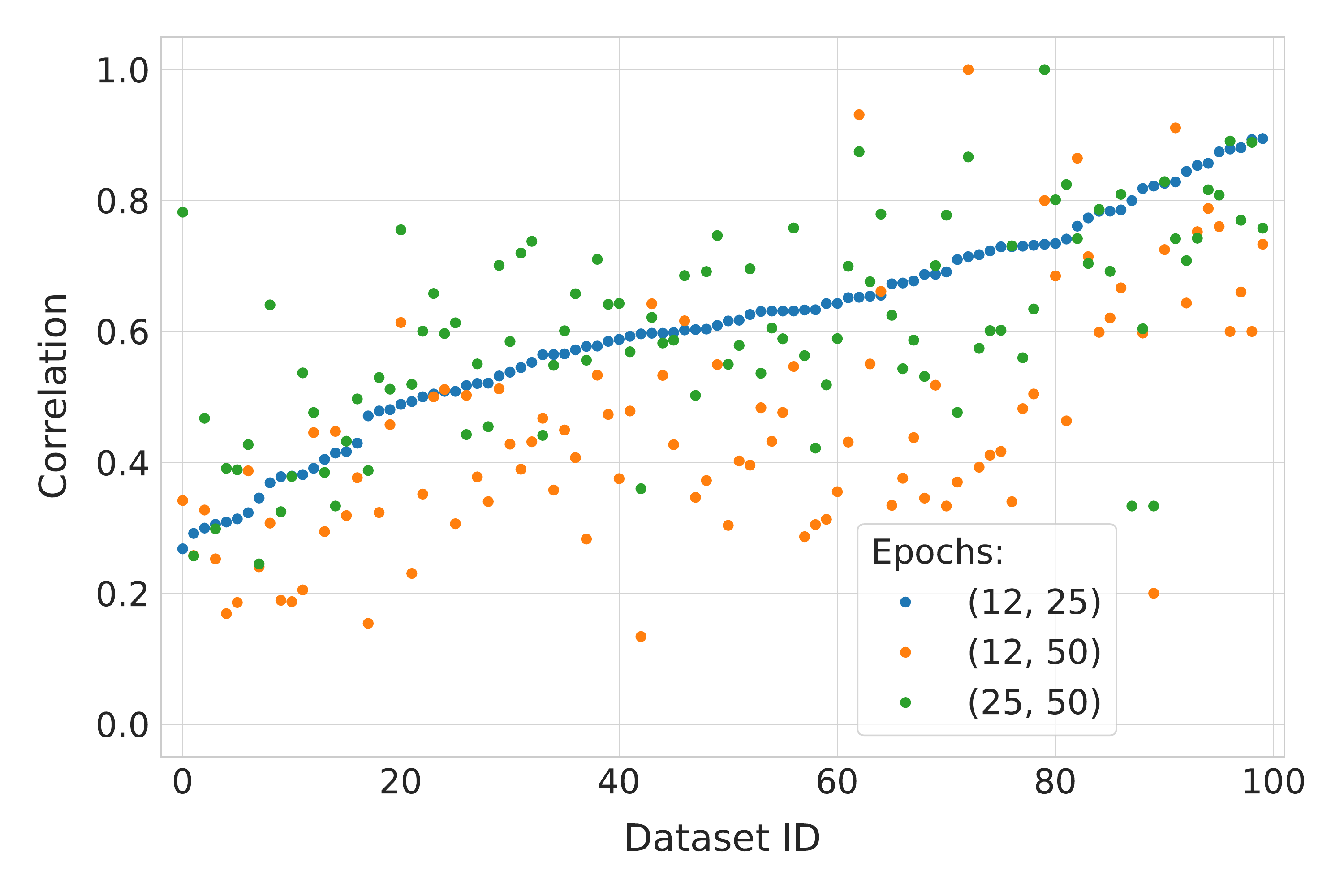}
    \label{fig:full_correlations}
\end{figure}

Figure~\ref{fig:full_correlations} shows the Spearman rank correlation between budgets for all meta training datasets. Whereas the correlation was consistently quite high on the small configuration space, the correlation is quite low on some datasets when using the larger configuration space. This indicates that there are specific hyperparameters in our larger configuration space whose settings are more sensitive to budgets and thus lead to smaller correlations.
Although the correlations are weaker compared to our study on LCBench, we will show in the upcoming experiments that the correlations suffice for Auto-PyTorch to perform well.

\begin{figure}[htp]
    \centering
    \caption{Average hyperparameter importance with fANOVA}
    \includegraphics[width=0.45\textwidth]{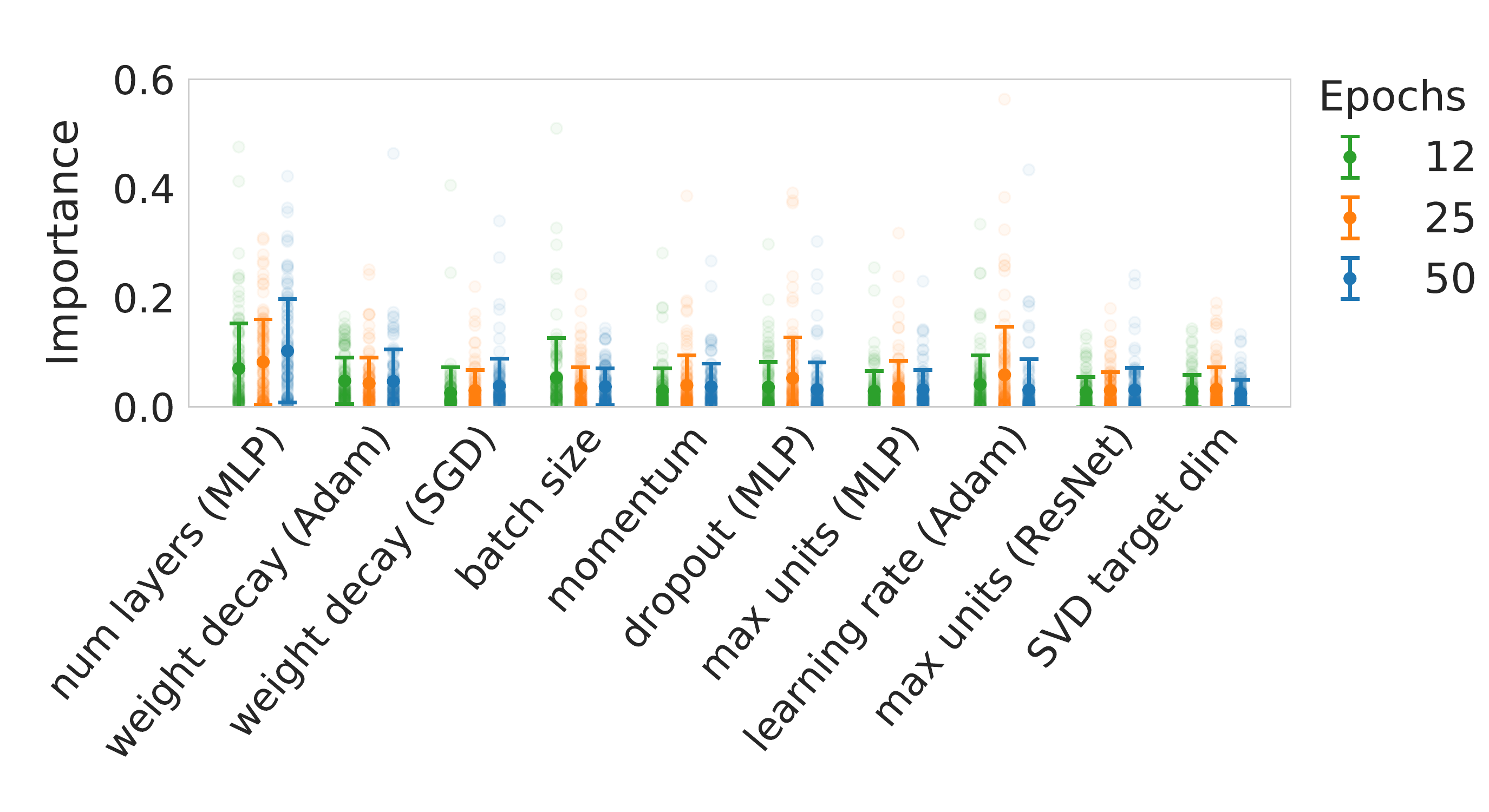}
    \label{fig:full_fanova}
\end{figure}

We also performed an fANOVA analysis and show the 10 most important hyperparameters in Figure~\ref{fig:full_fanova}. Overall, the importance is more evenly distributed than in the previous results.
Both architecture design and hyperparameters are both very important to be optimized, which indicates that both should be jointly optimized. Consistent with our previous results, the hyperparameter importance is quite stable across budgets, indicating that this is not the reason for the smaller correlations observed on some datasets.

\subsection{Ablation Results}

To understand the contribution of the individual components of Auto-PyTorch, we study its performance on the meta-test datasets by successively adding one component at a time. We start from plain Bayesian Optimization (BO) and move to multi-fidelity optimization with BOHB, then further add portfolios and finally ensembles. We show two datasets at a time in the following and refer to Figure~\ref{fig:all_datasets} in the appendix for all datasets and trajectories.

\subsubsection{Multi-fidelity Optimization}

\begin{figure}[htp]
    \centering
    \caption{Comparison of BO vs BOHB by showing the mean performance as solid lines and the standard deviation as shaded areas.}
    \includegraphics[width=0.45\textwidth]{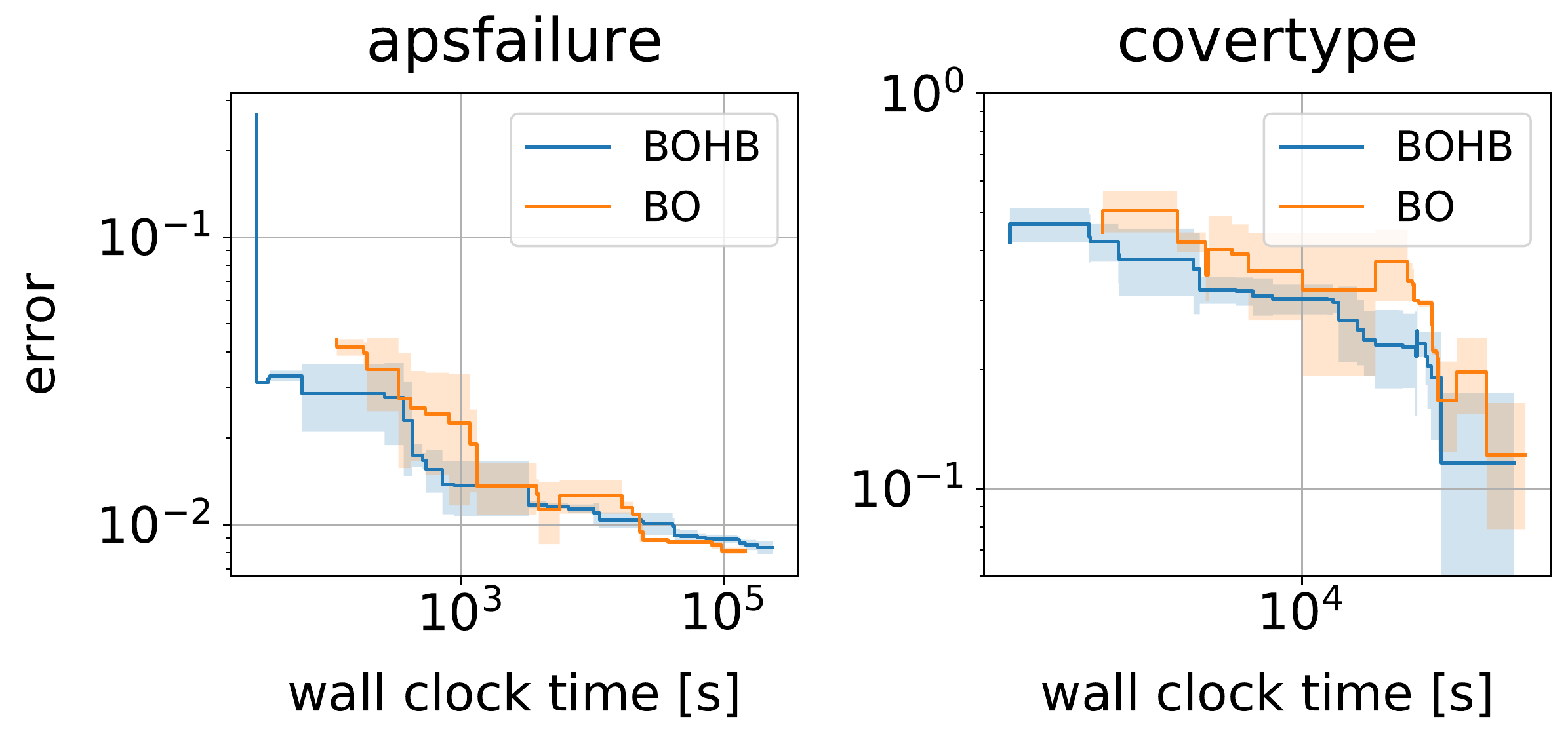}
    \label{fig:multi_fid}
\end{figure}

On all meta-test datasets, BOHB performed as well or better than plain BO in terms of anytime performance. In particular on the larger datasets, BOHB outperformed BO. Figure~\ref{fig:multi_fid} illustrates that BOHB can find a strong network earlier than BO by leveraging the lower budgets. Furthermore, BOHB manages to transfer to higher budgets without losing performance compared to BO even in the convergence limit.

\subsubsection{Portfolios}

\begin{figure}[htp]
    \centering
    \caption{Effect of adding portfolios for warmstarting the search. The solid lines show the mean performances and the shaded areas show the standard deviation.}
    \includegraphics[width=0.45\textwidth]{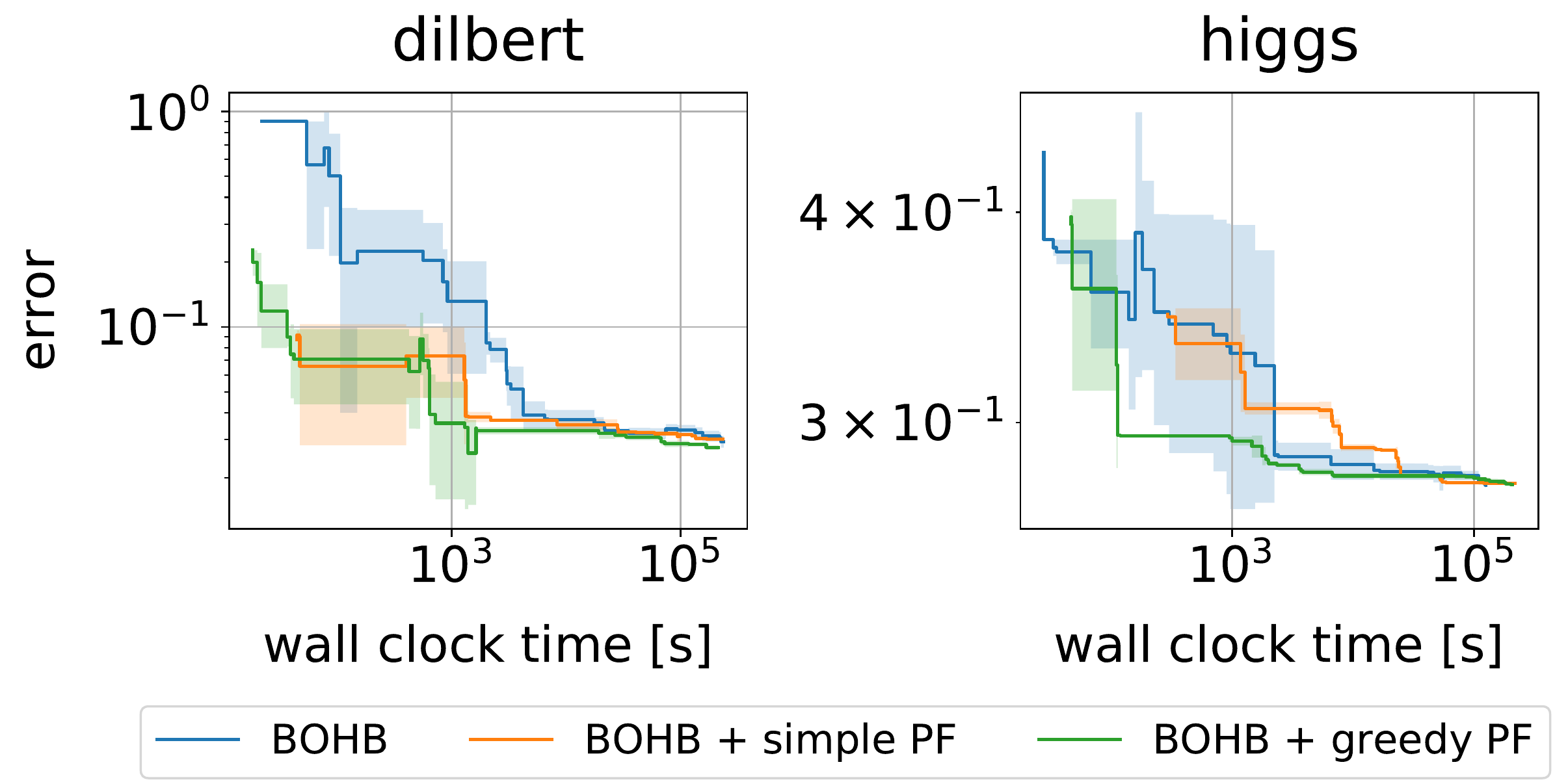}
    \label{fig:portfolios}
\end{figure}

Anytime performance further improves when we  add well-constructed portfolios to Auto-PyTorch. While the simple portfolio (with all incumbents of the $100$ meta-train datasets) sometimes performs worse than plain BOHB because of the induced overhead, the greedily constructed portfolio reliably improves performance (see Figure~\ref{fig:portfolios}). In particular, it matches the first few steps of BOHB and then shows a strong improvement once the first configurations are evaluated on the largest budget. This is expected, since the portfolio is constructed from configurations that perform well on the highest budget.

\subsubsection{Ensembles}

\begin{figure}[htp]
    \centering
    \caption{Comparison of different ensembles. The solid lines show the mean performances and the shaded areas refers the standard deviation. \textit{"BOHB $+$ greedy PF $+$ Ens. (div.)"} refers to the full Auto-PyTorch method.}
    \includegraphics[width=0.45\textwidth]{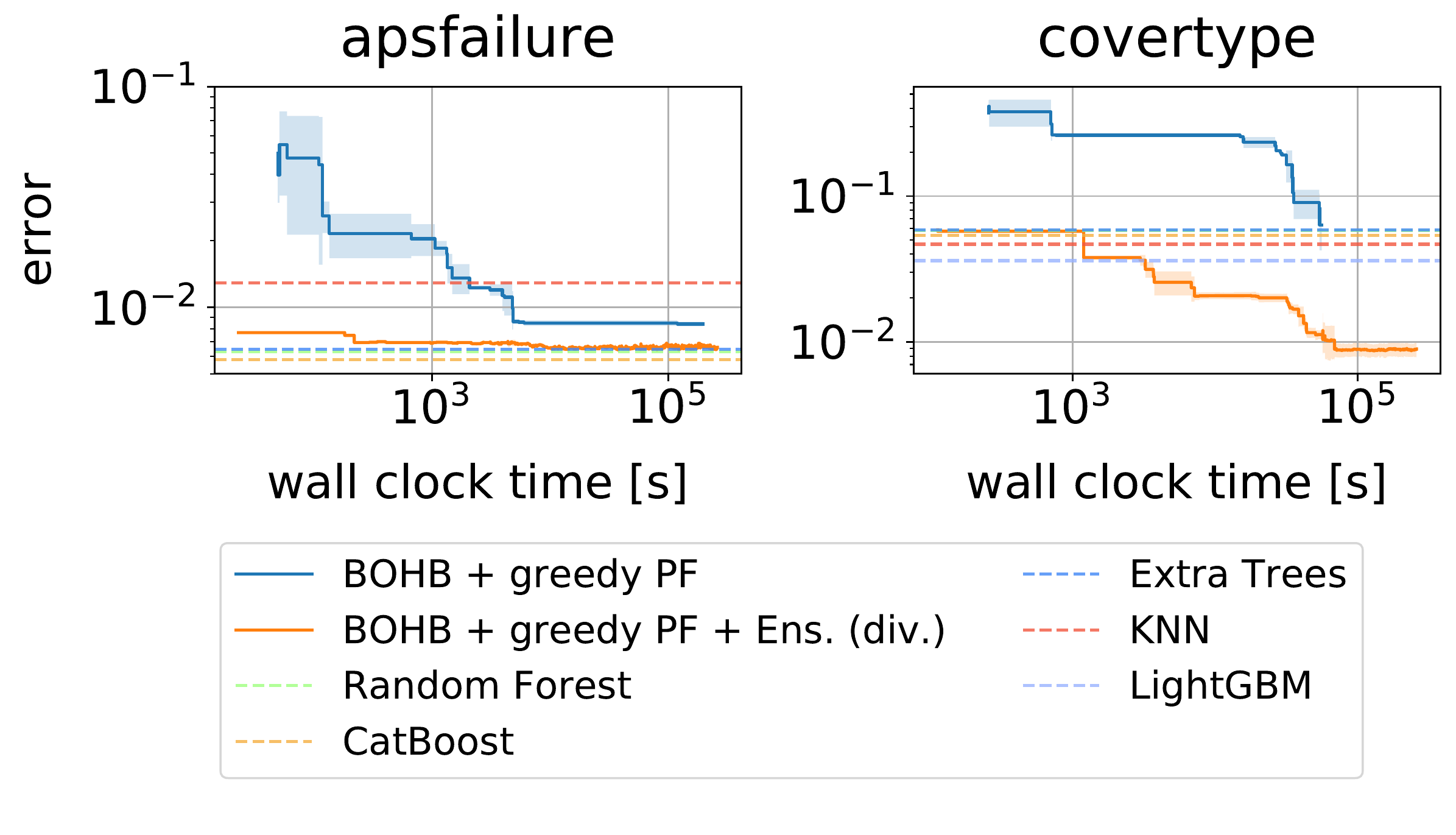}
    \label{fig:ensembles}
\end{figure}

Finally, we build ensembles from the evaluated configurations and compare to common baselines for tabular data.
Figure~\ref{fig:ensembles} shows that building ensembles from different DNNs and fidelities improves the performance in the long run, sometimes substantially. Furthermore, pure DL can outperform traditional baselines, but not all the time. 

To get the best of both worlds, we therefore propose to build diverse ensembles based on the trained DNNs and in addition based on the baselines, including Random Forest, Extra Trees, LightGBM, Catboost and KNN. Implementation-wise, we first train the baselines and start with the portfolio and BOHB afterwards.
By doing this, we can show that the robustness of Auto-PyTorch (i.e., BOHB + greedy portfolio + diverse ensembles) substantially improved s.t. it outperforms the baselines on all but one meta-test dataset. See Figure~\ref{fig:all_datasets} in the appendix.

\subsubsection{Parallelization}

\begin{figure}[htp]
    \centering
    \caption{BOHB with 3 parallel workers vs. sequential (not par.) BOHB}
    \includegraphics[width=0.45\textwidth]{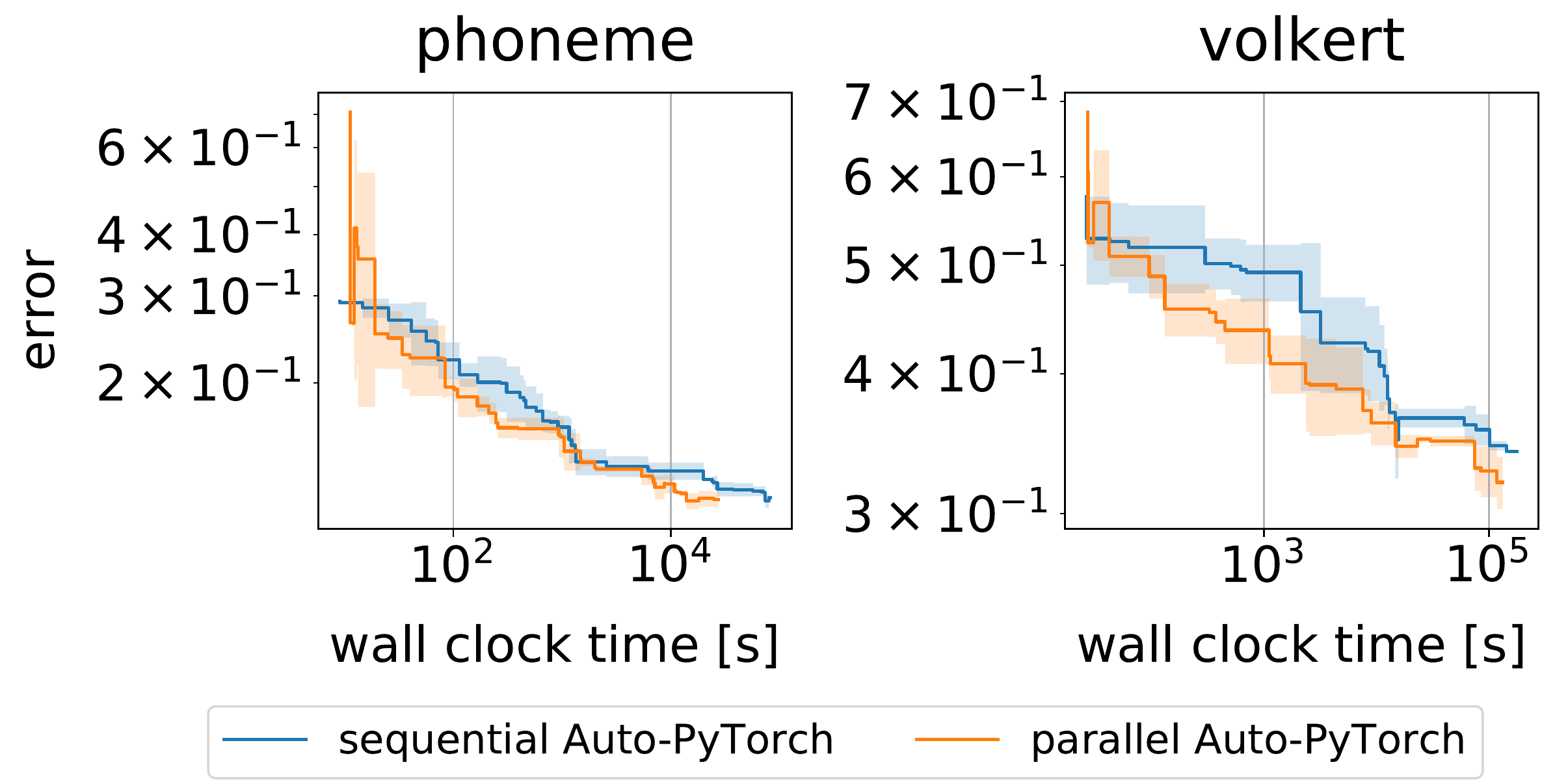}
    \label{fig:parallization}
\end{figure}

One of the advantages of Auto-PyTorch and its workhorse BOHB is that it nicely parallelizes to using  more available compute resources. In Figure~\ref{fig:parallization}, we show that parallel Auto-PyTorch makes efficient use of three parallel workers compared to its sequential version and can achieve speedups of $3$x or more in particular on larger datasets.

\subsection{Comparison against Common Baselines and Other AutoML Systems}

We compare Auto-PyTorch Tabular to several state-of-the-art AutoML frameworks, i.e. Auto-Keras~\cite{jin-kdd19a}, Auto-Sklearn\cite{feurer-nips2015a} and hyperopt-sklearn~\cite{komer-automl19a}. We also include the early version Auto-Net2.0~\cite{mendoza2019towards}. Whenever possible we chose a runtime of 1h. Alternatively, we ran with an increasing number of trials to obtain a trajectory and inferred the performance after 1h therefrom. The results are reported for 5~seeds in Table~\ref{tab:comparison} (left part). We note that some frameworks (AutoKeras and auto-sklearn) sometimes ran into memory errors and some (hyperopt-sklearn) could not handle missing data; all of these are denoted as ``-'' in the table.
After 1h, Auto-PyTorch Tabular clearly performed best overall. Indeed, based on a t-test with  $\alpha=0.05$, it performed statistically significantly better than each competitor on between 5 and 7 out of 7 datasets, only performing statistically significantly worse than auto-sklearn and hyperopt-sklearn on a single dataset. The largest absolute gain in accuracy that Auto-PyTorch Tabular achieved was for covertype, where it reduced the misclassification rate by a factor of 10 compared to the best competitor, AutoNet2.0: from $31.78\%$ to $3.14\%$.

\begin{table*}[tbh]
\centering
\caption{Accuracy and standard deviation across $5$ runs of Auto-PyTorch and $4$ previous AutoML frameworks after 1h. "-" indicates that the system crashed and did not return predictions. The best system in each column block and systems without sufficient evidence of being worse (using a t-test with a $p$-value of 0.05) are boldfaced. We further compare against the complementary and concurrently introduced approach of AutoGluon~\cite{erickson2020autogluon}, showing performance of both systems with and without ensembling and auto-stacking, respectively.}
\setlength{\tabcolsep}{3pt}
\begin{tabular}{@{}l|ccccc||cc||cc@{}}
\toprule
 & Auto-PyTorch & AutoNet2.0 & AutoKeras & Auto-Sklearn & hyperopt. & Auto-PyTorch & AutoGluon & Auto-PyTorch & AutoGluon\\
 & Tabular & \cite{mendoza2019towards} & \cite{jin-kdd19a}  & \cite{feurer-aaai15a} & \cite{komer-automl14a} & w/ ens. & w/ stack. & w/o ens. & w/o stack.\\

\midrule
\multicolumn{1}{l|}{covertype}   & \textbf{96.86 $\pm$ 0.41}                     & 68.22 $\pm$ 2.46                                                               & 61.61 $\pm$ 3.52                                                         & -                                                                           & -   & \textbf{96.86 $\pm$ 0.41} & \multicolumn{1}{c||}{-}                                          & \textbf{93.35 $\pm$ 0.02} & -                                          \\
\multicolumn{1}{l|}{volkert}     & \textbf{70.52 $\pm$ 0.51}                     & 60.90 $\pm$ 3.89                                                               & 44.25 $\pm$ 2.38                                                         & 67.32 $\pm$ 0.46                                                            & -    & 70.52 $\pm$ 0.51                           & \multicolumn{1}{c||}{\textbf{72.16 $\pm$ 0.00}}   & \textbf{70.74 $\pm$ 0.01} & 68.34 $\pm$ 0.10                                                                                                 \\
\multicolumn{1}{l|}{higgs}       & \textbf{73.01 $\pm$ 0.09}                     & 71.36 $\pm$ 0.55                                                               & 71.25 $\pm$ 0.29                                                         & 72.03 $\pm$ 0.33                                                            & -           & 73.01 $\pm$ 0.09                           & \multicolumn{1}{c||}{\textbf{73.26 $\pm$ 0.00}}   & \textbf{72.64 $\pm$ 0.02} & $72.60 \pm 0.00$                                                              \\
\multicolumn{1}{l|}{car}         & 96.41 $\pm$ 1.47                                               & 96.14 $\pm$ 0.35                                                               & 93.39 $\pm$ 2.82                                                         & \textbf{98.42 $\pm$ 0.62}                                  & \textbf{98.95 $\pm$ 0.96}   & $96.41 \pm 1.47$                           & \multicolumn{1}{c||}{\textbf{99.12 $\pm$ 0.50}}   & \textbf{98.59 $\pm$ 0.01} & $97.16 \pm 0.35$                             \\
\multicolumn{1}{l|}{mfeat-fact.} & \textbf{99.10 $\pm$ 0.18}                     & \textbf{98.97 $\pm$ 0.21}                                     & 97.73 $\pm$ 0.23                                                         & 98.64 $\pm$ 0.39                                                            & 97.88 $\pm$ 38.48             & \textbf{99.10 $\pm$ 0.18} & \multicolumn{1}{c||}{98.79 $\pm$ 0.15}                             & \textbf{98.78 $\pm$ 0.01} & $98.03 \pm 0.23$                                            \\
\multicolumn{1}{l|}{apsfailure}  & \textbf{99.41 $\pm$ 0.05}                     & 98.83 $\pm$ 0.03                                                               & -                                                                        & \textbf{99.43 $\pm$ 0.04}                                  & -                                     & 99.41 $\pm$ 0.05                           & \multicolumn{1}{c||}{\textbf{99.57 $\pm$ 0.01}}   & 99.38 $\pm$ 0.00                           & \textbf{99.50 $\pm$ 0.03}                                         \\
\multicolumn{1}{l|}{phoneme}     & \textbf{90.55 $\pm$ 0.14}                     & 86.61 $\pm$ 0.19                                                               & 86.76 $\pm$ 0.12                                                         & 89.26 $\pm$ 0.14                                                            & 89.79 $\pm$ 4.54                        & \textbf{90.55 $\pm$ 0.14} & \multicolumn{1}{c||}{\textbf{90.63 $\pm$ 0.08}}   & \textbf{90.40 $\pm$ 0.01} & $89.62 \pm 0.06$                                                          \\
\multicolumn{1}{l|}{dilbert}                          & \textbf{98.70 $\pm$ 0.15}                     & 97.43 $\pm$ 0.46                                                               & 96.51 $\pm$ 0.62                                                         & 98.14 $\pm$ 0.47                                                            & -      & 98.70 $\pm$ 0.15                           & \multicolumn{1}{c||}{\textbf{99.33 $\pm$ 0.00}}   & \textbf{98.54 $\pm$ 0.00} & $98.17 \pm 0.05$                                                                             \\ \bottomrule
\end{tabular}
\label{tab:comparison}
\end{table*}

We also compared Auto-PyTorch to the complementary and concurrently introduced AutoGluon~\cite{erickson2020autogluon} which uses automated stacking to build ensembles. To study the benefits of ensembles, we show results of Auto-PyTorch with and without ensembles alongside AutoGluon with and without auto-stacking in Table~\ref{tab:comparison} (two right parts). Auto-PyTorch performed overall better than Auto-Gluon without ensembles and stacking (which is the default of the AutoGluon software), showing that the search procedure and space of Auto-PyTorch work well. For both systems, ensembling boosts their performance significantly. Auto-Gluon's auto-stacking provides a small edge above Auto-PyTorch, leading us to expect that employing stacking ensembles in Auto-PyTorch could further improve its overall performance.

\subsection{Image Classification}

\begin{figure}[htb!]
    \centering
    \includegraphics[width=0.3\textwidth]{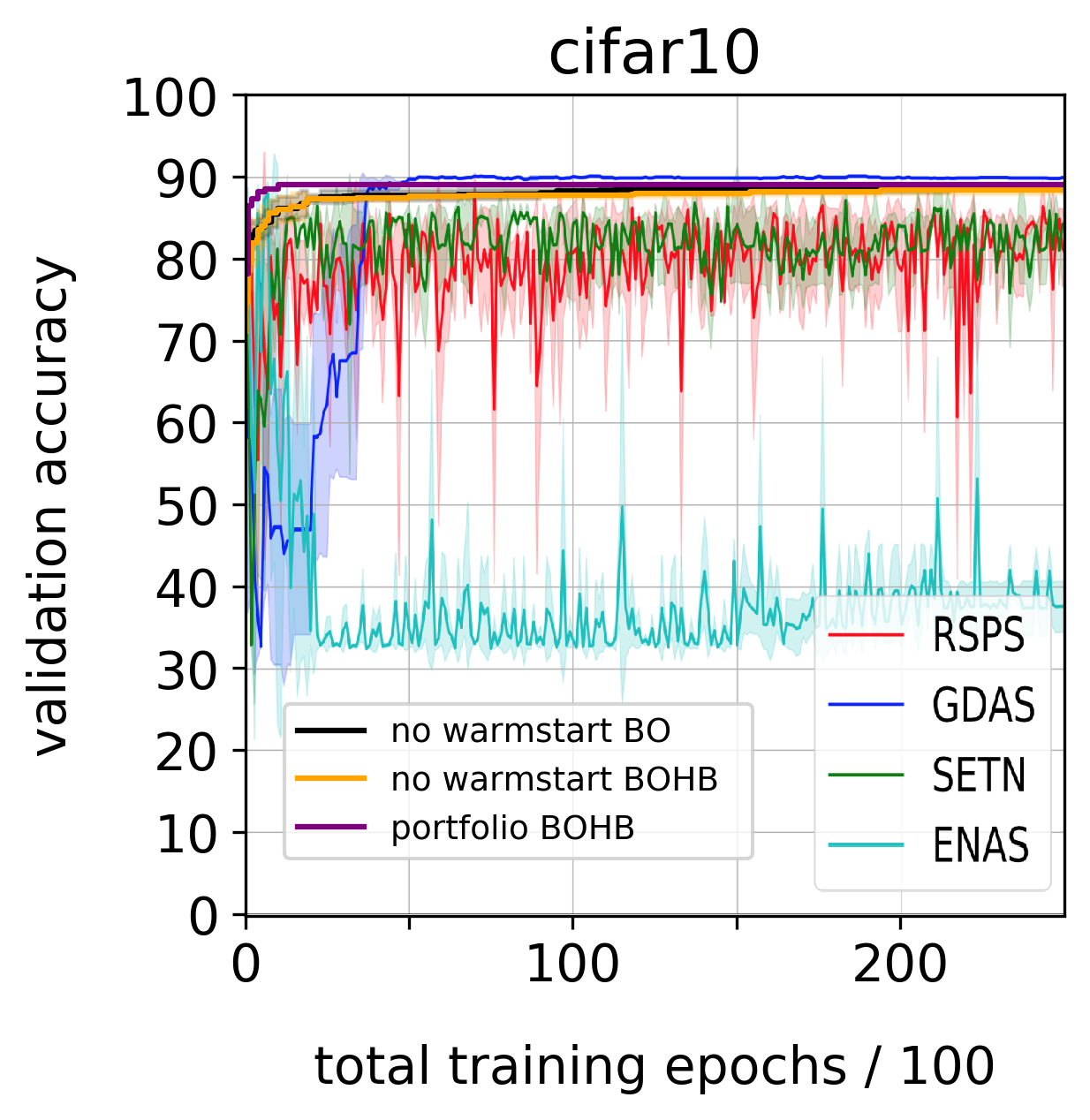}
    \caption{Performance of Auto-PyTorch on \mbox{NAS-Bench-201} as compared to different one-shot optimizers.}
    \label{fig:nb201_oneshot}
\end{figure}

To show that Auto-PyTorch's methodological core of multi-fidelity meta-learning is also applicable to other data modalities, we performed a proof-of-concept experiment using NAS-Bench-201~\cite{dong-iclr20a}, a NAS benchmark for object recognition. We ran Auto-PyTorch (without ensembles and portfolio) on the \mbox{NAS-Bench-201} search space for 500 iterations, using 25, 50 and 100 epochs as budgets, for all three datasets provided by \mbox{NAS-Bench-201} (CIFAR10, CIFAR100, and downsampled ImageNet-16-120). We then again ran Auto-PyTorch on each dataset, this time using a portfolio consisting of the 5 best configurations found on either of the remaining two datasets.
We compare our results to the results of different one-shot optimizers as reported by the authors of \mbox{NAS-Bench-201}~\cite{dong-iclr20a} in Figure~\ref{fig:nb201_oneshot}. Auto-PyTorch (``portfolio BOHB'') exhibits a very strong anytime performance and outperforms all one-shot methods except for GDAS~\cite{dong2019searching}. We also find that using a portfolio outperforms plain BOHB. Note that we did not evaluate the benefit of ensembles as \mbox{NAS-Bench-201} does not provide the trained models or predictions. We further study the benefit of multi-fidelity optimization and portfolios in an ablation study on all three datasets of NAS-Bench-201 in Appendix~\ref{app:image_data}.

\section{Discussion \& Conclusion}\label{sec:discussion}

In this paper, we introduced Auto-PyTorch Tabular, a robust approach for jointly optimizing architectures and hyperparameters of deep neural networks. To make educated design decisions, we first studied the challenges of the task on a newly proposed learning curve benchmark, dubbed LCBench. Although LCBench only covers a small configuration space, most of the insights gained on it actually transferred to the more complex configuration space of Auto-PyTorch Tabular. We are currently in the process of extending LCBench to larger datasets and more fidelities.

Our main insights can be summarized as: (i) a well-designed configuration space for joint architecture search and hyperparameter optimization allows to find very well-performing models for a wide range of datasets, (ii) multi-fidelity optimization combined with Bayesian optimization can efficiently search in such spaces, (iii) complementary portfolios found on meta-train datasets improve the performance substantially, in particular in the early phase of the optimization, (iv) by constructing ensembles out of the trained DNNs and strong baselines for tabular data, Auto-PyTorch Tabular achieves strong performance and outperforms several state-of-the-art competitors, and (v) Auto-PyTorch's methodological multi-fidelity meta-learning core also performs well for object recognition in NAS-Bench-201.

\section{Future Works}
\label{sec:future_works}
Although Auto-PyTorch performs well, there are several ways to push it still further. First of all, while the ensembles improve performance overall, it is also known that they can lead to overfitting. By using better generalization performance estimates~\cite{dwork2015reusable,tsamardinos-ml2018a}, the performance of our framework could therefore be improved. Furthermore, while we currently use default hyperparameter settings for all ensemble members other than neural networks, a combination of Auto-PyTorch with Auto-Sklearn~\cite{feurer-nips2015a,feurer-arxiv20a} could lead to an even stronger performance. Also, we studied the importance of hyperparameters but have not yet made use of the results to improve Auto-PyTorch's performance further. 
Last but not least, while we have mainly demonstrated Auto-PyTorch in the domain of tabular data, its underlying methodology is not specific to any data modality, and in future work we aim to apply it to many more tasks involving image, text and time series data.

\ifCLASSOPTIONcompsoc
  \section*{Acknowledgments}
\else
  \section*{Acknowledgment}
\fi

The authors acknowledge funding by the Robert Bosch GmbH and support by the European Research Council (ERC) under the European Unions Horizon 2020 research and innovation program through grant no. 716721.

\bibliography{strings,lib,local,procE}

\begin{thebibliography}{10}

\bibitem{automl_book}
F.~Hutter, L.~Kotthoff, and J.~Vanschoren, editors.
\newblock {\em Automatic Machine Learning: Methods, Systems, Challenges}.
\newblock Challenges in Machine Learning. Springer, 2019.

\bibitem{thornton-kdd13a}
C.~Thornton, F.~Hutter, H.~Hoos, and K.~Leyton-Brown.
\newblock {A}uto-{WEKA}: combined selection and hyperparameter optimization of
  classification algorithms.
\newblock In {\em The 19th {ACM} {SIGKDD} International Conference on Knowledge
  Discovery and Data Mining ({KDD}'13)}, 2013.

\bibitem{komer-automl14a}
B.~Komer, J.~Bergstra, and C.~Eliasmith.
\newblock Hyperopt-sklearn: Automatic hyperparameter configuration for
  scikit-learn.
\newblock In {\em {ICML} workshop on Automated Machine Learning (Auto{ML}
  2014)}, 2014.

\bibitem{olson-gecco16a}
R.~Olson, N.~Bartley, R.~Urbanowicz, and J.~Moore.
\newblock Evaluation of a {Tree}-based {Pipeline} {Optimization} {Tool} for
  {Automating} {Data} {Science}.
\newblock In {\em Proceedings of the Genetic and Evolutionary Computation
  Conference ({GECCO}'16)}, pages 485--492, 2016.

\bibitem{feurer-aaai15a}
M.~Feurer, T.~Springenberg, and F.~Hutter.
\newblock Initializing {B}ayesian hyperparameter optimization via
  meta-learning.
\newblock In {\em Proceedings of the Twenty-nineth National Conference on
  Artificial Intelligence ({AAAI}'15)}, pages 1128--1135. {AAAI} Press, 2015.

\bibitem{jin-kdd19a}
H.~Jin, Q.~Song, and X.Hu.
\newblock Auto-keras: An efficient neural architecture search system.
\newblock In {\em Proceedings of the 25th {ACM} {SIGKDD} International
  Conference on Knowledge Discovery {\&} Data Mining, {KDD} 2019}, pages
  1946--1956, 2019.

\bibitem{zoph-iclr17a}
B.~Zoph and Q.~V. Le.
\newblock Neural architecture search with reinforcement learning.
\newblock In {\em Proceedings of the international conference on representation
  learning {(ICLR)}}, 2017.

\bibitem{elsken-jmlr19a}
T.~Elsken, J.~Metzen, and F.~Hutter.
\newblock Neural architecture search: {A} survey.
\newblock {\em Journal of Machine Learning Research}, 20:55:1--55:21, 2019.

\bibitem{zela-automl18a}
A~Zela, A.~Klein, S.~Falkner, and F.~Hutter.
\newblock Towards automated deep learning: Efficient joint neural architecture
  and hyperparameter search.
\newblock In {\em ICML 2018 AutoML Workshop}, 2018.

\bibitem{ying-icml19a}
C.~Ying, A.~Klein, E.~Christiansen, E.~Real, K.~Murphy, and F.~Hutter.
\newblock {NAS-Bench-101:} towards reproducible neural architecture search.
\newblock In {\em Proceedings of the 36th International Conference on Machine
  Learning ({ICML}'19)}, volume~97, pages 7105--7114. Proceedings of Machine
  Learning Research, 2019.

\bibitem{yang-iclr20a}
A.~Yang, P.~Esperana, and F.~Carlucci.
\newblock {NAS} evaluation is frustratingly hard.
\newblock In {\em Proceedings of the International Conference on Learning
  Representations ({ICLR})}. OpenReview.net, 2020.

\bibitem{mendoza-automl16a}
H.~Mendoza, A.~Klein, M.~Feurer, J.~Springenberg, and F.~Hutter.
\newblock Towards automatically-tuned neural networks.
\newblock In {\em ICML 2016 AutoML Workshop}, 2016.

\bibitem{guyon-ijcnn15a}
I.~Guyon, K.~Bennett, G.~Cawley, H.~J. Escalante, S.~Escalera, Tin~Kam Ho,
  N.~Macià, B.~Ray, M.~Saeed, A.~Statnikov, and E.~Viegas.
\newblock Design of the 2015 chalearn automl challenge.
\newblock In {\em 2015 International Joint Conference on Neural Networks
  ({IJCNN})}, pages 1--8. IEEE Computer Society Press, 2015.

\bibitem{falkner-icml18a}
S.~Falkner, A.~Klein, and F.~Hutter.
\newblock {BOHB}: {R}obust and {E}fficient {H}yperparameter {O}ptimization at
  {S}cale.
\newblock In {\em Proceedings of the international conference on machine
  learning ({ICML})}, pages 1437--1446, 2018.

\bibitem{feurer-automl18a}
M.~Feurer, K.~Eggensperger, S.~Falkner, M.~Lindauer, and F.~Hutter.
\newblock Practical automated machine learning for the automl challenge 2018.
\newblock In {\em AutoML workshop at international conference on machine
  learning (ICML)}, 2018.

\bibitem{dong-iclr20a}
X.~Dong and Y.~Yang.
\newblock {NAS-Bench-201:} extending the scope of reproducible neural
  architecture search.
\newblock In {\em Proceedings of the International Conference on Learning
  Representations ({ICLR})}. OpenReview.net, 2020.

\bibitem{automl-book-hpo}
M.~Feurer and F.~Hutter.
\newblock Hyperparameter optimization.
\newblock In Hutter et~al. \cite{automl_book}, pages 3--38.

\bibitem{stanley2002evolving}
K.~O. Stanley and R.~Miikkulainen.
\newblock Evolving neural networks through augmenting topologies.
\newblock {\em Evolutionary computation}, 10(2):99--127, 2002.

\bibitem{pham-icml18a}
H.~Pham, M.~Guan, B.~Zoph, Q.~Le, and J.~Dean.
\newblock Efficient neural architecture search via parameter sharing.
\newblock In {\em Proceedings of the international conference on machine
  learning ({ICML})}, pages 4092--4101, 2018.

\bibitem{liu-iclr19}
H.~Liu, K.~Simonyan, and Y.~Yang.
\newblock {DARTS:} differentiable architecture search.
\newblock In {\em Proceedings of the International Conference on Learning
  Representations ({ICLR}'19)}, 2019.
\newblock Published online: \url{iclr.cc}.

\bibitem{zela-iclr20a}
A.~Zela, T.~Elsken, T.~Saikia, Y.~Marrakchi, T.~Brox, and F.~Hutter.
\newblock Understanding and robustifying differentiable architecture search.
\newblock In {\em Proceedings of the International Conference on Learning
  Representations ({ICLR})}. OpenReview.net, 2020.

\bibitem{mendoza2019towards}
H.~Mendoza, A.~Klein, M.~Feurer, J.~T. Springenberg, M.~Urban, M.~Burkart,
  M.~Dippel, M.~Lindauer, and F.~Hutter.
\newblock Towards automatically-tuned deep neural networks.
\newblock In Hutter et~al. \cite{automl_book}, pages 135--149.

\bibitem{lindauer-arxiv19a}
M.~Lindauer and F.~Hutter.
\newblock Best practices for scientific research on neural architecture search.
\newblock {\em Journal of Machine Learning Research}, 21:1--21, 2020.

\bibitem{eggensperger-bayesopt13}
K.~Eggensperger, M.~Feurer, F.~Hutter, J.~Bergstra, J.~Snoek, H.~Hoos, and
  K.~Leyton-Brown.
\newblock Towards an empirical foundation for assessing {Bayesian} optimization
  of hyperparameters.
\newblock In {\em NIPS Workshop on {B}ayesian Optimization in Theory and
  Practice (BayesOpt'13)}, 2013.

\bibitem{bischl-aij16a}
B.~Bischl, P.~Kerschke, L.~Kotthoff, M.~Lindauer, Y.~Malitsky,
  A.~Frech\'{e}tte, H.~Hoos, F.~Hutter, K.~Leyton-Brown, K.~Tierney, and
  J.~Vanschoren.
\newblock {ASlib}: A benchmark library for algorithm selection.
\newblock {\em Artificial Intelligence}, pages 41--58, 2016.

\bibitem{zela-iclr20b}
A.~Zela, J.~Siems, and F.~Hutter.
\newblock {NAS-Bench-1Shot1:} benchmarking and dissecting one-shot neural
  architecture search.
\newblock In {\em Proceedings of the International Conference on Learning
  Representations ({ICLR})}. OpenReview.net, 2020.

\bibitem{sharma-dis19a}
A.~Sharma, J.~van Rijn, F.~Hutter, and A.~M{\"{u}}ller.
\newblock Hyperparameter importance for image classification by residual neural
  networks.
\newblock In {\em Proceedings of the international conference on Discovery
  Science {DS}}, volume 11828, pages 112--126. Springer, 2019.

\bibitem{rijn-kdd18a}
J.~van Rijn and F.~Hutter.
\newblock Hyperparameter importance across datasets.
\newblock In {\em Proceedings of the 24th {ACM} {SIGKDD} International
  Conference on Knowledge Discovery and Data Mining ({KDD})}, pages 2367--2376.
  ACM Press, 2018.

\bibitem{hutter-icml14a}
F.~Hutter, H.~Hoos, and K.~Leyton-Brown.
\newblock An efficient approach for assessing hyperparameter importance.
\newblock In {\em Proceedings of the 31th International Conference on Machine
  Learning, ({ICML}'14)}, 2014.

\bibitem{biedenkapp-lion18a}
A.~Biedenkapp, J.~Marben, M.~Lindauer, and F.~Hutter.
\newblock {CAVE}: Configuration assessment, visualization and evaluation.
\newblock In {\em Proceedings of the International Conference on Learning and
  Intelligent Optimization ({LION})}, Lecture Notes in Computer Science.
  Springer, 2018.

\bibitem{pytorch}
A.~Paszke, S.~Gross, S.~Chintala, G.~Chanan, E.~Yang, Z.~DeVito, Z.~Lin,
  A.~Desmaison, L.~Antiga, and A.~Lerer.
\newblock Automatic differentiation in {PyTorch}.
\newblock In {\em {N}eur{IPS} Autodiff Workshop}, 2017.

\bibitem{lindauer2019a}
M.~Lindauer, K.~Eggensperger, M.~Feurer, A.~Biedenkapp, J.~Marben, P.~Müller,
  and F.~Hutter.
\newblock {BOAH}: A tool suite for multi-fidelity bayesian optimization {\&}
  analysis of hyperparameters.
\newblock {\em arXiv:1908.06756 {[cs.LG]}}, 2019.

\bibitem{kotila-autonomio}
M.~Kotila.
\newblock Autonomio.
\newblock \url{https://mikkokotila.github.io/slate/}, 2019.

\bibitem{srivastava-jmlr14a}
N.~Srivastava, G.~Hinton, A.~Krizhevsky, I.~Sutskever, and R.~Salakhutdinov.
\newblock Dropout: A simple way to prevent neural networks from overfitting.
\newblock {\em Journal of Machine Learning Research}, 15:1929--1958, 2014.

\bibitem{kingma-iclr15a}
D.~Kingma and J.~Ba.
\newblock Adam: {A} method for stochastic optimization.
\newblock In {\em Proceedings of the International Conference on Learning
  Representations ({ICLR}'15)}, 2015.

\bibitem{zhang2017-mixup}
H.~Zhang, M.~Ciss{\'{e}}, Y.~Dauphin, and D.~Lopez{-}Paz.
\newblock mixup: Beyond empirical risk minimization.
\newblock In {\em Proceedings of the International Conference on Learning
  Representations ({ICLR}'18)}, 2018.
\newblock Published online: \url{iclr.cc}.

\bibitem{gastaldi2017-shakeshake}
X.~Gastaldi.
\newblock Shake-shake regularization of 3-branch residual networks.
\newblock In {\em 5th International Conference on Learning Representations,
  Workshop Track Proceedings}. OpenReview.net, 2017.

\bibitem{yamada2018-shakedrop}
Y.~Yamada, M.~Iwamura, T.~Akiba, and K.~Kise.
\newblock Shakedrop regularization for deep residual learning.
\newblock {\em {IEEE} Access}, 7:186126--186136, 2019.

\bibitem{he2015-resnet}
K.~He, X.~Zhang, S.~Ren, and J.~Sun.
\newblock Deep residual learning for image recognition.
\newblock In {\em Proceedings of the {IEEE} Conference on Computer Vision and
  Pattern Recognition, ({CVPR})}, pages 770--778. {IEEE} Computer Society,
  2016.

\bibitem{mockus-jgo94}
J.~Mockus.
\newblock Application of {B}ayesian approach to numerical methods of global and
  stochastic optimization.
\newblock {\em Journal of Global Optimization}, 4(4):347--365, 1994.

\bibitem{li-jmlr18a}
L.~Li, K.~Jamieson, G.~DeSalvo, A.~Rostamizadeh, and A.~Talwalkar.
\newblock Hyperband: A novel bandit-based approach to hyperparameter
  optimization.
\newblock {\em Journal of Machine Learning Research}, 18(185):1--52, 2018.

\bibitem{jamieson-aistats16}
K.~Jamieson and A.~Talwalkar.
\newblock Non-stochastic best arm identification and hyperparameter
  optimization.
\newblock In {\em Proceedings of the International Conference on Artificial
  Intelligence and Statistics ({AISTATS})}, volume~51, 2016.

\bibitem{klein-aistats17}
A.~Klein, S.~Falkner, S.~Bartels, P.~Hennig, and F.~Hutter.
\newblock Fast {Bayesian} optimization of machine learning hyperparameters on
  large datasets.
\newblock In {\em Proceedings of the International Conference on Artificial
  Intelligence and Statistics ({AISTATS})}, volume~54, 2017.

\bibitem{scikit-learn}
F.~Pedregosa, G.~Varoquaux, A.~Gramfort, V.~Michel, B.~Thirion, O.~Grisel,
  M.~Blondel, P.~Prettenhofer, R.~Weiss, V.~Dubourg, J.~Vanderplas, A.~Passos,
  D.~Cournapeau, M.~Brucher, M.~Perrot, and E.~Duchesnay.
\newblock Scikit-learn: Machine learning in {P}ython.
\newblock {\em Journal of Machine Learning Research}, 12:2825--2830, 2011.

\bibitem{feurer-nips2015a}
M.~Feurer, A.~Klein, K.~Eggensperger, J.~T. Springenberg, M.~Blum, and
  F.~Hutter.
\newblock Efficient and robust automated machine learning.
\newblock In {\em Proceedings of the 29th International Conference on Advances
  in Neural Information Processing Systems ({N}eur{IPS}'15)}, pages 2962--2970,
  2015.

\bibitem{caruana-icml04a}
R.~Caruana, A.~Niculescu-Mizil, G.~Crew, and A.~Ksikes.
\newblock Ensemble selection from libraries of models.
\newblock In {\em Proceedings of the 21st International Conference on Machine
  Learning ({ICML}'04)}. Omnipress, 2004.

\bibitem{caruana-icdm06a}
R.~Caruana, A.~Munson, and A.~Niculescu-Mizil.
\newblock Getting the most out of ensemble selection.
\newblock In {\em Proceedings of the 6th {IEEE} International Conference on
  Data Mining ({ICDM}'06)}, pages 828--833. IEEE Computer Society Press, 2006.

\bibitem{xu-aaai10a}
L.~Xu, H.~Hoos, and K.~Leyton-Brown.
\newblock Hydra: Automatically configuring algorithms for portfolio-based
  selection.
\newblock In {\em Proceedings of the Twenty-fourth National Conference on
  Artificial Intelligence ({AAAI}'10)}, pages 210--216. {AAAI} Press, 2010.

\bibitem{xu-rcra11a}
L.~Xu, F.~Hutter, H.~Hoos, and K.~Leyton-Brown.
\newblock {Hydra-MIP}: Automated algorithm configuration and selection for
  mixed integer programming.
\newblock In {\em Proc. of RCRA workshop at IJCAI}, 2011.

\bibitem{vanschoren-sigkdd14a}
J.~Vanschoren, J.~van Rijn, B.~Bischl, and L.~Torgo.
\newblock {OpenML}: Networked science in machine learning.
\newblock {\em SIGKDD Explorations}, 15(2):49--60, 2014.

\bibitem{gijsbers-automl}
P.~Gijsbers, E.~LeDell, J.~Thomas, S.~Poirier, B.~Bischl, and J.~Vanschoren.
\newblock An open source {AutoML} benchmark.
\newblock In {\em {ICML AutoML} Workshop}, 2019.

\bibitem{loshchilov-iclr17a}
I.~Loshchilov and F.~Hutter.
\newblock Sgdr: Stochastic gradient descent with warm restarts.
\newblock In {\em Proceedings of the international conference on representation
  learning {(ICLR)}}, 2017.

\bibitem{erickson2020autogluon}
N.~Erickson, J.~Mueller, A.~Shirkov, H.~Zhang, P.~Larroy, M.~Li, and A.~Smola.
\newblock {AutoGluon-Tabular}: Robust and accurate automl for structured data.
\newblock {\em CoRR}, abs/2003.06505, 2020.

\bibitem{ke2017-lightgbm}
G.~Ke, Q.~Meng, T.~Finley, T.~Wang, W.~Chen, W.~Ma, Q.~Ye, and T.~Liu.
\newblock {LightGBM:} {A} highly efficient gradient boosting decision tree.
\newblock In {\em Proceedings of the 31st International Conference on Advances
  in Neural Information Processing Systems ({N}eur{IPS}'17)}, 2017.

\bibitem{dorogush2018-catboost}
A.~Dorogush, V.~Ershov, and A.~Gulin.
\newblock {CatBoost:} gradient boosting with categorical features support.
\newblock {\em CoRR}, abs/1810.11363, 2018.

\bibitem{breimann-mlj01a}
L.~Breimann.
\newblock Random forests.
\newblock {\em Machine Learning Journal}, 45:5--32, 2001.

\bibitem{geurts_ml2006a}
P.~Geurts, D.~Ernst, and L.~Wehenkel.
\newblock Extremely randomized trees.
\newblock {\em Machine Learning}, 63(1):3--42, 2006.

\bibitem{komer-automl19a}
B.~Komer, J.~Bergstra, and C.~Eliasmith.
\newblock Hyperopt-sklearn.
\newblock In Hutter et~al. \cite{automl_book}, pages 97--111.

\bibitem{dong2019searching}
X.~Dong and Y.~Yang.
\newblock Searching for a robust neural architecture in four gpu hours.
\newblock In {\em Proceedings of the IEEE Conference on computer vision and
  pattern recognition}, pages 1761--1770, 2019.

\bibitem{dwork2015reusable}
C.~Dwork, V.~Feldman, M.~Hardt, T.~Pitassi, O.~Reingold, and A.~Roth.
\newblock The reusable holdout: Preserving validity in adaptive data analysis.
\newblock {\em Science}, 349(6248):636--638, 2015.

\bibitem{tsamardinos-ml2018a}
I.~Tsamardinos, E.~Greasidou, and G.~Borboudakis.
\newblock Bootstrapping the out-of-sample predictions for efficient and
  accurate cross-validation.
\newblock {\em Machine Learning}, 107(12):1895--1922, 2018.

\bibitem{feurer-arxiv20a}
M.~Feurer, K.~Eggensperger, S.~Falkner, M.~Lindauer, and F.~Hutter.
\newblock Auto-sklearn 2.0: The next generation.
\newblock {\em arXiv:2007:04074 [cs.LG]}, 2020.

\bibitem{feurer2019openml-pythonapi}
M.~Feurer, J.~N. van Rijn, A.~Kadra, P.~Gijsbers, N.~Mallik, S.~Ravi,
  A.~M{\"u}ller, J.~Vanschoren, and F.~Hutter.
\newblock {OpenML-Python: an extensible Python API for OpenML}.
\newblock {\em arXiv:1911.02490}, 2019.

\end{thebibliography}
\bibliographystyle{unsrt}

\clearpage
\appendices

\section{Datasets for Portfolio Construction}
\label{app:datasets}

\begin{figure}[htp]
    \centering
    \caption{Distribution of meta-train datasets ("meta") and meta-test datasets. Meta-train datasets are further categorized by datasets used in Section~\ref{sec:exploration} ("LCBench") and additional datasets ("new")}
    \includegraphics[width=0.48\textwidth]{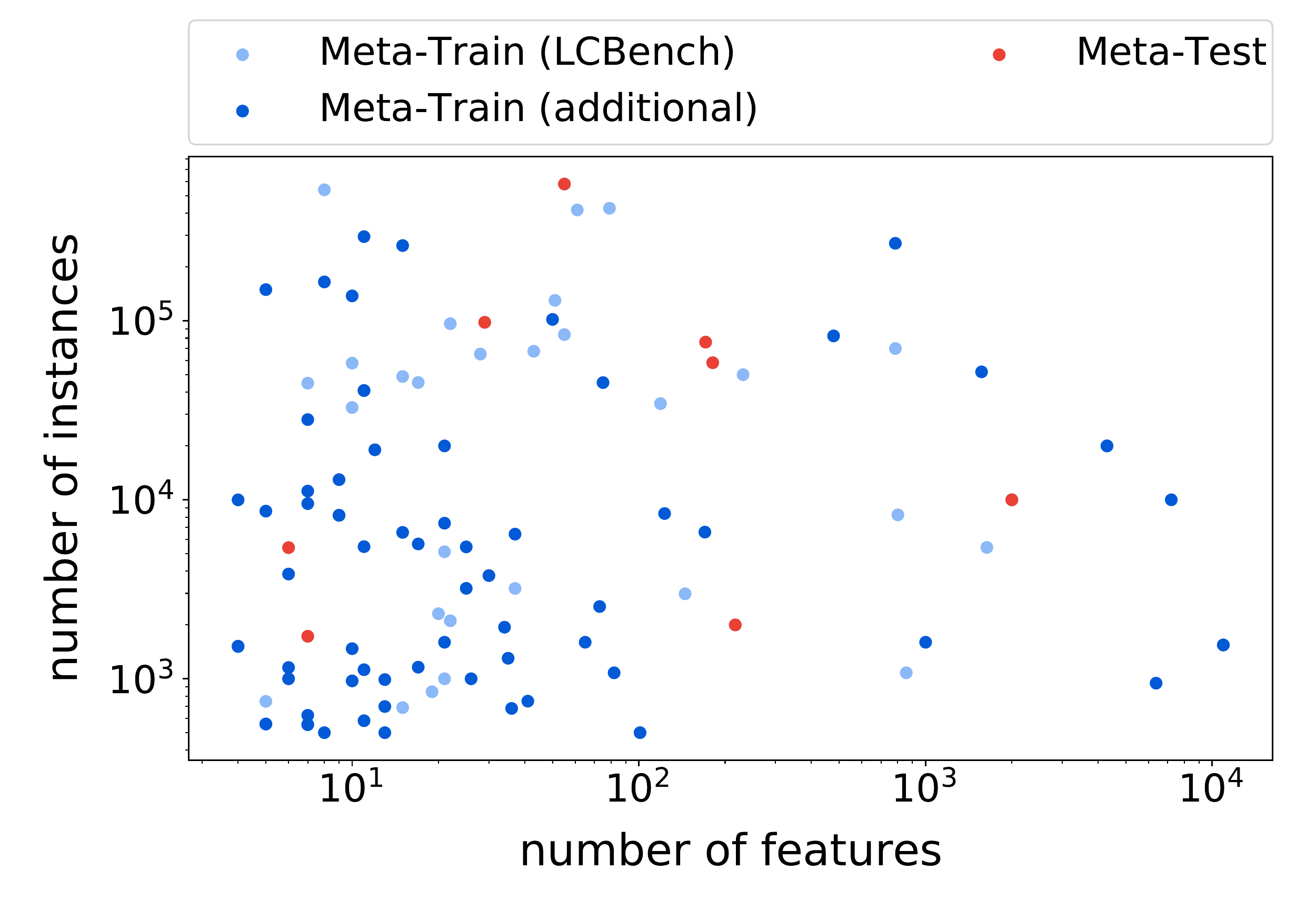}
    \label{fig:feat_vs_inst}
\end{figure}

The datasets used for constructing a portfolio were obtained by sampling datasets from clusters of similar metafeatres (e.g. number of instances, classes) using the OpenML-Python API~\cite{feurer2019openml-pythonapi}. Only datasets with at least two attributes and between $500$ and $1\,000\,000$ data points were sampled. Sparse and synthetic datasets, as well as datasets containing time or string type attributes were not considered. Finally, we checked for overlaps between $\mathbf{D}_{\text{meta}}$ and $\mathbf{D}_{\text{test}}$. Figure~\ref{fig:feat_vs_inst} visualizes the distribution of datasets with regards the number of instances and features. The datasets cover a large range of these metafeatures.

\section{Ablation Trajectories on all Meta-test Datasets}

\begin{figure*}[htb!]
    \centering
    \includegraphics[width=0.98\textwidth]{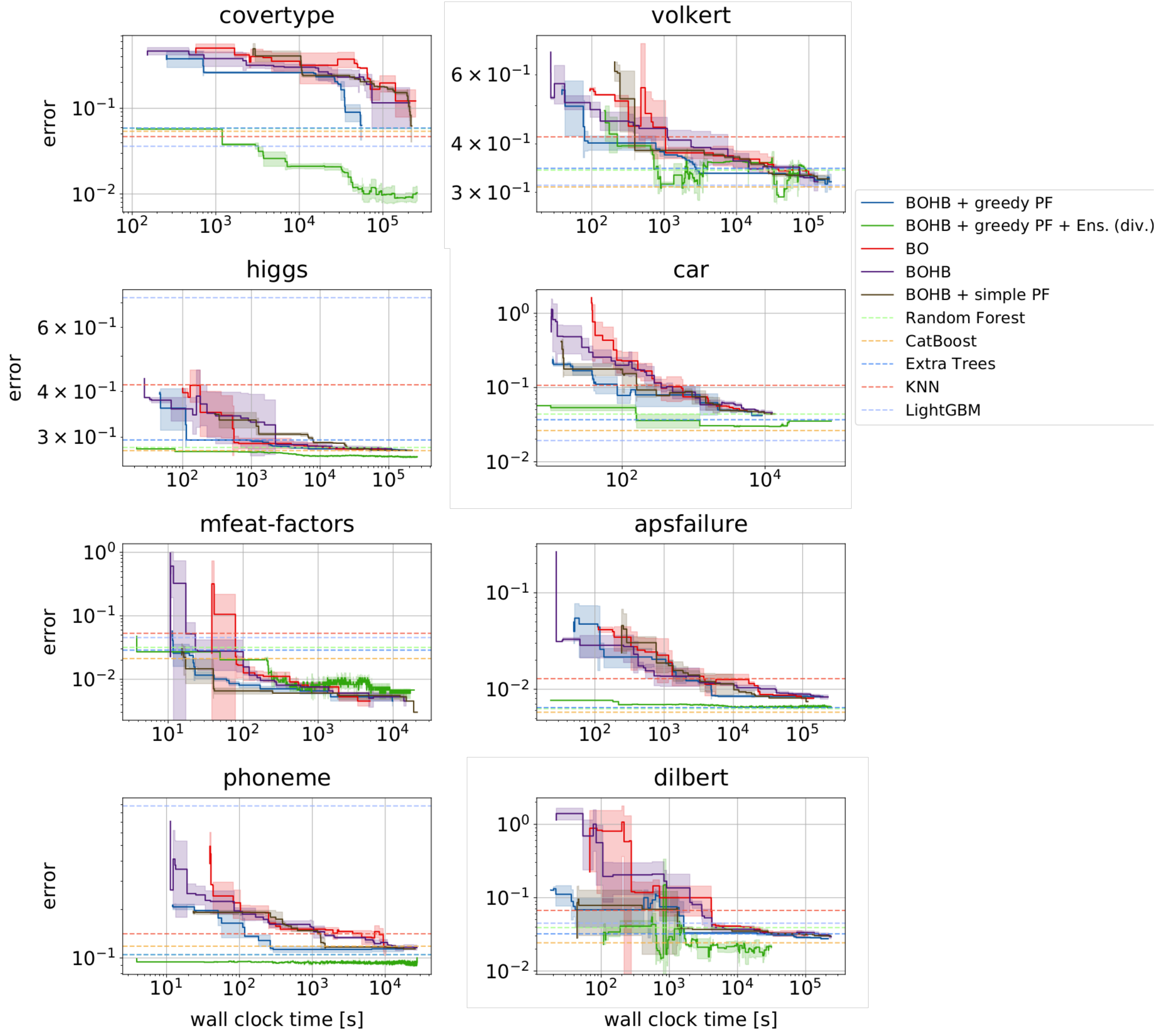}
    \caption{Results on all meta-test datasets. \textit{"BOHB $+$ greedy PF $+$ Ens. (div.)"} refers to the full Auto-PyTorch method.}
    \label{fig:all_datasets}
\end{figure*}

Figure~\ref{fig:all_datasets} shows the results of the ablations study on all meta-test datasets. For the baselines we used scikit-sklearn version 0.23.0 (Random Forest, Extra Trees, KNN), LightGBM version 2.3.1 and Catboost version 0.23.1. 

\section{Comparison against Common Baselines and AutoML Frameworks}

\begin{figure*}[htb!]
    \centering
    \includegraphics[width=0.98\textwidth]{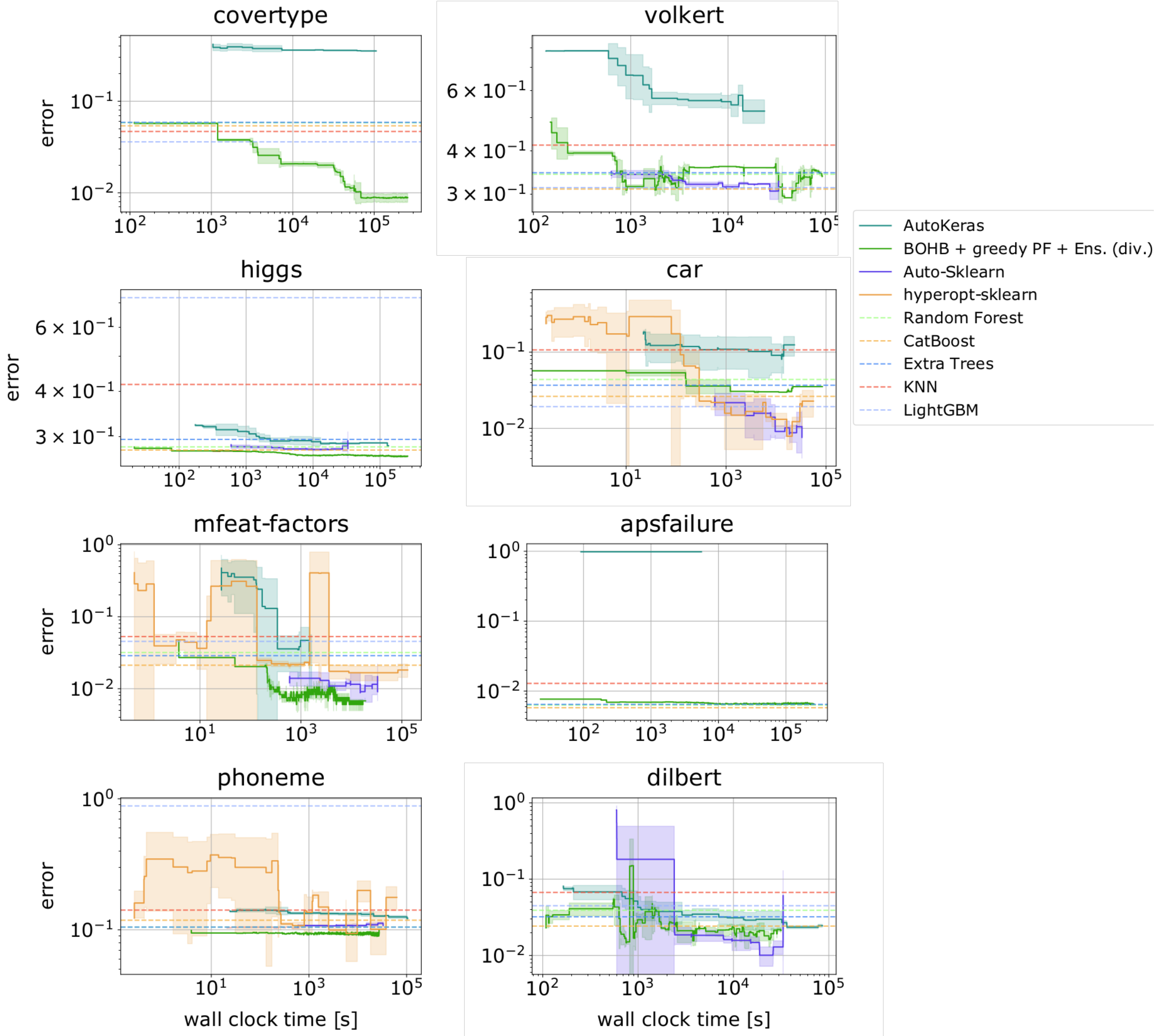}
    \caption{Comparison against competitors on all meta-test datasets. \textit{"BOHB $+$ greedy PF $+$ Ens. (div.)"} refers to the full Auto-PyTorch method.}
    \label{fig:all_datasets_comparison}
\end{figure*}

Figure~\ref{fig:all_datasets_comparison} shows the results of Auto-PyTorch and other common AutoML frameworks as well as our baseline models on all meta-test datasets. For all experiments, the same resources were allocated (i.e. 6 Intel Xeon Gold 6242 CPU cores, 6 GB RAM per core). For Auto-PyTorch the logging post-hoc ensemble selection allows the construction of the full trajectory. For other frameworks, the trajectories were created by running for $1, 3, 5, 10, 30, 50, ..., 10\,000$ trials and using the timestamp of the trial finish as timestamp for the point on the trajectory.

We used Auto-Keras 1.0.2 and the default settings. For AutoGluon, we used version 0.0.9 in the "optimize for deployment" setting with the same validation split as Auto-PyTorch uses. For hyperopt-sklearn, we allowed any classifier and preprocessing to be used and used the tpe alogrithm. Finally, we used auto-sklearn 0.7.0 with an ensemble size of 50, initializing with the 50 best, model selection via $33~\%$ holdout, 3 workers. All other settings were set to their default value.

\section{Proof of Concept Experiment with Image Data}
\label{app:image_data}

\begin{figure*}[htb!]
    \centering
    \includegraphics[width=0.98\textwidth]{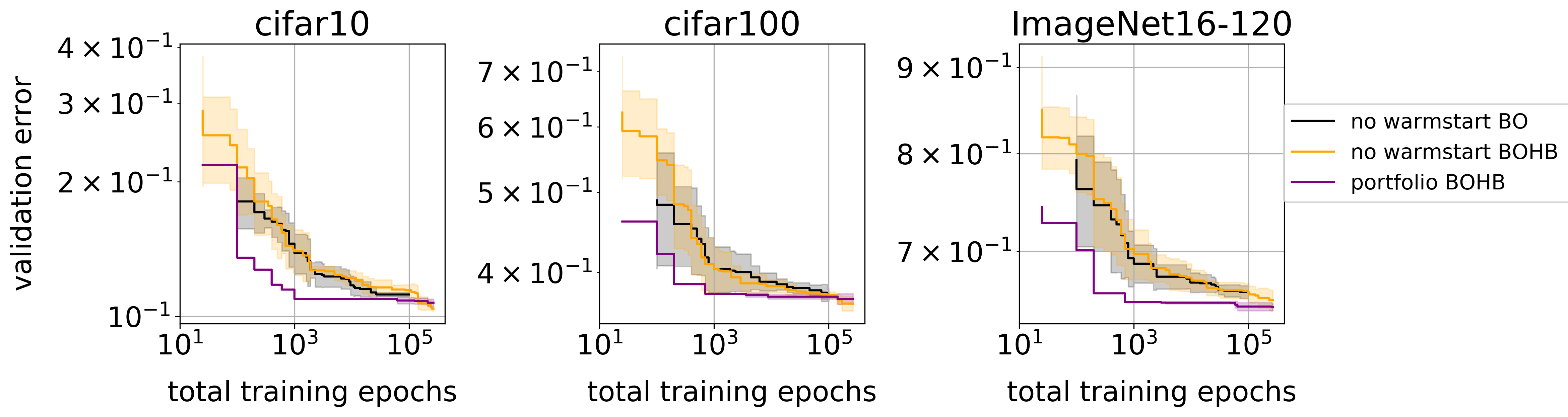}
    \caption{Performance of Auto-PyTorch on \mbox{NAS-Bench-201} only using BO, using BOHB and using BOHB plus a portfolio.}
    \label{fig:nb201_ablation}
\end{figure*}

Figure~\ref{fig:nb201_ablation} shows the performance of Auto-PyTorch when ablating multi-fidelity optimization and portfolios on all three datasets. We find that BOHB yields faster solutions compared to BO with similar performance after long runtimes. Most notably, the introduction of a portfolio improves anytime performance on all datasets significantly.

\end{document}